\documentclass[sigconf, 10pt]{acmart}
\usepackage{todonotes}
\usepackage{graphicx}
\usepackage{subcaption} %
\usepackage{url}
\usepackage{hyperref}
\usepackage{tikz}
\usepackage{amsmath}
\usepackage{enumitem}

\newcommand*\circledsmall[1]{\tikz[baseline=(char.base)]{
            \node[shape=circle,draw,inner sep=1pt] (char) {#1};}}

\AtBeginDocument{%
  }

 \setcopyright{none}
\renewcommand\footnotetextcopyrightpermission[1]{} %

\begin{document}

\title{LinguaLinked: A Distributed Large Language Model Inference System for Mobile Devices}

\author{Junchen Zhao}
\authornote{These authors contributed equally to this research.}
\email{junchez3@uci.edu}
\affiliation{%
  \institution{University of California Irvine}
  \city{Irvine}
  \state{CA}
  \country{USA}
  \postcode{92617}
}

\author{Yurun Song}
\authornotemark[1]
\email{yuruns@uci.edu}
\affiliation{%
  \institution{University of California Irvine}
  \city{Irvine}
  \state{CA}
  \country{USA}
  \postcode{92617}
}

\author{Simeng Liu}
\authornotemark[1]
\email{simenl3@uci.edu}
\affiliation{%
  \institution{University of California Irvine}
  \city{Irvine}
  \state{CA}
  \country{USA}
  \postcode{92617}
}

\author{Ian G. Harris}
\email{harris@ics.uci.edu}
\affiliation{%
  \institution{University of California Irvine}
  \city{Irvine}
  \state{CA}
  \country{USA}
  \postcode{92617}
}

\author{Sangeetha Abdu Jyothi}
\email{sangeetha.aj@uci.edu}
\affiliation{%
 \institution{University of California Irvine, VMware Research}
  \city{Irvine}
  \state{CA}
  \country{USA}
  \postcode{92617}
}

\renewcommand{\shortauthors}{Anonymous}

\begin{abstract}
  Deploying Large Language Models (LLMs) locally on mobile devices presents a significant challenge due to their extensive memory requirements. In this paper, we introduce LinguaLinked, a system for decentralized, distributed LLM inference on mobile devices. LinguaLinked enables collaborative execution of the inference task across multiple trusted devices. LinguaLinked ensures data privacy by processing information locally. LinguaLinked uses three key strategies. First, an optimized model assignment technique segments LLMs and uses linear optimization to align segments with each device's capabilities. Second, an optimized data transmission mechanism ensures efficient and structured data flow between model segments while also maintaining the integrity of the original model structure. Finally, LinguaLinked incorporates a runtime load balancer that actively monitors and redistributes tasks among mobile devices to prevent bottlenecks, enhancing the system's overall efficiency and responsiveness. We demonstrate that LinguaLinked facilitates efficient LLM inference while maintaining consistent throughput and minimal latency through extensive testing across various mobile devices, from high-end to low-end Android devices. In our evaluations, compared to the baseline, LinguaLinked achieves an inference performance acceleration of $1.11\times$ to $1.61\times$ in single-threaded settings, $1.73\times$ to $2.65\times$ with multi-threading. Additionally, runtime load balancing yields an overall inference acceleration of $1.29\times$ to $1.32\times$.
\end{abstract}

\maketitle
\pagestyle{plain}

\section{Introduction}
The past decade has witnessed a seismic shift in the machine learning (ML) landscape, particularly with the rise of large language models (LLMs), which are built atop transformer decoders \cite{vaswani2023attention}. These LLMs [\citealp{brown2020language}, \citealp{kaplan2020scaling}, \citealp{hoffmann2022training}, \citealp{chowdhery2022palm}, \citealp{zhang2022opt}, \citealp{touvron2023llama}, \citealp{workshop2023bloom}] have achieved state-of-art performance on Natural Language Processing (NLP) benchmarks such as text generation, question answering, machine translation, and text summarization, and led to commercial offerings such as OpenAI ChatGPT \cite{Introduc51:online} and Github Copilot \cite{GitHubCo19:online}. Recent research has established that as the number of parameters in these models increases, they demonstrate enhanced capabilities in various language tasks [\citealp{alabdulmohsin2022revisiting}, \citealp{Clark2022UnifiedSL}, \citealp{10.1145/3373376.3378530}, \citealp{patel2022mapping}, \citealp{hendrycks2021measuring}, \citealp{cobbe2021training}].

Due to their large sizes, LLMs are usually deployed on servers for inference serving. For instance, a Vicuna-13B \cite{peng2023instruction} with full precision requires over 52GB runtime memory and cannot be loaded in highly resource-constrained devices. However, server-based inference poses several challenges: (1) sending data to servers poses a privacy concern, as sensitive information may be exposed during transmission or at the server itself; (2) centralized computation typically requires high bandwidth for communication, which can be costly and inefficient, particularly for users in areas with limited Internet connectivity.

Mobile computing exemplifies a scenario with such challenges. By design, mobile devices process data at the source or close to it, alleviating bandwidth requirements and also mitigating privacy concerns. Yet, mobile devices frequently have limited processing capabilities and memory, making the accommodation of LLMs, with their high memory demands, especially challenging. To tackle the difficulties of deploying LLMs on resource-limited mobile devices, one strategy is to apply aggressive weight quantization \cite{mlc-llm}. By reducing the precision of LLM weights to 3-bit or 4-bit levels, memory consumption can be substantially decreased. However, quantization may lead to a trade-off, potentially sacrificing model accuracy to enhance performance and reduce the memory requirement. Moreover, some LLMs may not be accommodated in mobile devices even after aggressive quantization.

An alternative approach is the distributed deployment of LLMs across multiple mobile devices. By partitioning the LLM into smaller segments, each segment can be allocated to different devices, and the inference can be executed across devices. Such distribution can alleviate the need for drastic reductions in precision. As a result, the LLM can maintain a higher level of accuracy while still being feasibly deployed within the resource constraints of mobile computing environments. While previous studies have looked into distributed model deployment on mobile computing platforms [\citealp{hu2019deephome}, \citealp{naveen2021low}, \citealp{coedge}, \citealp{adaptive_para_exec}], these have largely concentrated on smaller-scale models used in computer vision applications, which have a much smaller memory footprint compared to LLMs and typically do not need iterative inference.

In this paper, we present \textit{LinguaLinked}, a decentralized distributed inference system for LLM deployment on mobile devices. The core concept behind \textit{LinguaLinked} is to distribute segments of an LLM across multiple mobile devices, which then work together to serve inference queries. This approach allows us to operate within the constraints of mobile devices, which typically have limited processing power and memory. In LinguaLinked, each mobile device is tasked with only a fraction of the total computation and storage, thereby reducing the burden on any single device and allowing the collective network of devices to handle complex tasks collaboratively. Importantly, in our system, these devices are considered 'trusted.' For instance, an individual’s personal devices (e.g., a smartphone and a tablet) or shared mobile devices among a group of trusted contacts of family/friends can collaboratively host different segments of the LLM, ensuring both efficiency and a layer of privacy and security. Such segmentation also reduces the need for heavy quantization of model weights. 

While the distributed deployment of LLMs on mobile devices offers many benefits, it also presents several challenges:\\
(1) \textit{Model Assignment in a Heterogenous Environment.} In mobile computing environments, which are typically heterogeneous in nature, model segments of an LLM should be assigned to diverse devices in alignment with the varied computational and memory capabilities of individual devices. Effective model assignment must consider both the constraints of each device and the inter-dependencies between the layers of the LLMs. \\
(2) \textit{Residual Data Dependencies.} Residual data dependency refers to the reliance of one model segment on outputs from other model segments that are not adjacent to it. For example, the model segment on device 4 might depend on model segment residual output from device 1. These dependencies can significantly complicate the inference process, as they require a coordinated and often synchronous exchange of intermediate data across the network. \\
(3) \textit{Task Redistribution in a Dynamic Environment.} Mobile devices have fluctuating workloads and network conditions. The available resources on mobile devices may change during runtime, such as the available memory and bandwidth. An overloaded device may become a bottleneck in the inference process, which would lead to reduced system performance or failure. Therefore, it is important to redistribute tasks in runtime to optimize resource utilization and avoid overburdening any single device. 

\textit{LinguaLinked} tackles these challenges with three key system components: \\
(1) \textit{\textbf{Optimized Model Assignment}.} The optimized model assignment process starts with segmenting the LLM into smaller model segments. This segmentation is done independently of the state of mobile devices, focusing on minimizing model segment inter-dependencies and data output. Subsequently, linear optimization is employed for grouping the model segments and assigning them to appropriate devices while taking into account the hardware and network conditions of the mobile devices. This approach ensures the model segments assigned to each mobile device align with the device capabilities while reducing data transmission overhead between devices.\\
(2) \textit{\textbf{Runtime Load Balancing}.} The runtime load balancing strategy is capable of monitoring the system in runtime, identifying potential bottlenecks, and redistributing model segments accordingly. It takes into account the computational load, network conditions, and available resources on each device, then reassigns tasks from an overloaded device to underutilized ones, thus maintaining the overall efficiency and robustness of the system.\\ 
(3) \textit{\textbf{Optimized Communication}.} \textit{LinguaLinked} creates a set of data transmission maps to facilitate the flow of data between model segments on different devices. This approach ensures that data outputs from each segment are transferred effectively in accordance with these maps while maintaining the integrity of the original model structure. Each device uses these maps to determine the precise timing and destination for sending or receiving data, thereby optimizing the latency of data transmission.

We perform a thorough evaluation of \textit{LinguaLinked} on high-end and low-end Android devices. In a single-threaded setting, compared to the baseline, \textit{LinguaLinked} achieves an inference performance acceleration of approximately $1.11\times$ to $1.61\times$ across both quantized and full-precision models. With multi-threading, the system exhibits further improvements, achieving acceleration rates of approximately $1.73\times$ to $2.65\times$ for both quantized and full-precision models. Runtime load balancing yields an overall inference acceleration of $1.29\times$ to $1.32\times$. Importantly, our findings indicate that LinguaLinked's performance gains are more pronounced with larger models, suggesting enhanced scalability and effectiveness in handling complex, resource-intensive tasks. We develop an Android application to demonstrate LinguaLinked's effectiveness in a typical mobile computing environment, showing how LLMs can operate on devices with diverse capabilities.

The major contributions of our work are:
\begin{itemize}[leftmargin=*,nolistsep]
    \item We develop \textit{LinguaLinked}, the first system for decentralized distributed LLM inference on mobile devices.
    \item We design an optimized model assignment strategy that employs linear optimization for device-specific assignment, effectively aligning with mobile device capabilities and reducing data transmission overhead.
    \item We build a runtime load balancer that actively monitors and redistributes tasks among mobile devices to prevent bottlenecks and optimize overall system performance.
    \item We implement an optimized data transmission mechanism that utilizes data transmission maps to facilitate efficient and structured data flow between model segments.
    \item Our system enables inference on mobile devices for LLMs that do not fit in a single device and offers up to 2.65$\times$ acceleration compared to the baseline. 
\end{itemize}

\section{Background and Motivation}
\subsection{Background}

\subsubsection{Challenges in LLM Inference with Autoregressive Decoders}
Autoregressive decoder-based LLMs represent a significant advancement in the field of NLP. These models, such as GPT-3 \cite{brown2020language}, OPT \cite{zhang2022opt}, and LLaMA \cite{touvron2023llama}, operate by predicting one word at a time in a sequential manner, building subsequent output based on the previously generated words. This method allows for the generation of coherent and contextually relevant text, making these models highly effective for tasks such as language generation, translation, and conversation. However, the sequential nature of these models can pose challenges in terms of computational efficiency and latency, particularly when generating longer text passages, as each token must be predicted based on the entire preceding context [\citealp{lin2021limitations}, \citealp{gpt3-limit}, \citealp{math11112451}]. 

Traditionally, the processing of these models has been centralized in data centers equipped with powerful servers [\citealp{aminabadi2022deepspeed}, \citealp{borzunov2022petals}, \citealp{du2023improving}]. Based on this architecture, recent studies focused on the development, evaluation, and enhancement of LLM inference performance mostly on centralized servers. Tabi \cite{wang2023tabi} reduces inference latency by building a multi-level inference system. INFaaS \cite{romero2021infaas} builds an automated model-less system for distributed inference serving. Cocktail \cite{gunasekaran2022cocktail} proposes a
cost-effective model serving system in the public cloud, reducing deployment cost and inference latency. While this setup offers significant computational power, it raises concerns about privacy, as data must travel from the user's device to a central server, increasing the risk of data exposure [\citealp{khowaja2023chatgpt}, \citealp{sebastian2023chatgpt}, \citealp{renaud2023chatgpt}, \citealp{kshetri2023cybercrime}]. Additionally, centralized processing can result in latency, especially when users are geographically distant from servers or in situations of network congestion [\citealp{elbamby2019wireless}, \citealp{liang2020toward}, \citealp{park2019wireless}, \citealp{mao2017survey}]. 

\subsubsection{Mobile Device Constraints in LLM Deployment}
Mobile computing offers significant benefits such as reduced latency, lower bandwidth usage, and enhanced privacy, making it an increasingly popular choice for deploying models, including LLMs  [\citealp{wu2019machine}, \citealp{zhao2022survey}, \citealp{chen2019deep}, \citealp{zhang2019deep}]. However, the deployment of LLMs on mobile devices is challenging due to their limited computational and memory capacities. Common approaches to mitigate these challenges include model quantization [\citealp{gholami2022survey}, \citealp{bondarenko2021understanding}, \citealp{coelho2021automatic}], distillation [\citealp{liang2020mixkd}, \citealp{gu2023knowledge}, \citealp{jiao2019tinybert}], and pruning [\citealp{blalock2020state}, \citealp{hoefler2021sparsity}, \citealp{liang2021pruning}]. While these techniques are effective in reducing the size of the model, they can affect model performance and output quality, particularly when applied more aggressively [\citealp{gholami2022survey}, \citealp{guo2018survey}]. 

Frameworks like TensorFlow Lite \cite{TensorFlowLite}, TVM\cite{chen2018tvm}, and ONNXRuntime \cite{ONNXRuntime} have played a crucial role in facilitating the execution of models on mobile and embedded devices. Complementing these frameworks, recent advances in quantization techniques [\citealp{yao2022zeroquant}, \citealp{frantar2022gptq}, \citealp{xiao2023smoothquant}] have made it possible to deploy LLMs more efficiently. These methods manage to maintain accuracy while significantly reducing the computational time and memory requirements of the models. However, despite these improvements, the deployment of LLMs on mobile devices, particularly on low-end devices, remains challenging. This is primarily due to their limited memory capacity, which becomes a more pressing issue when the model's size exceeds the memory available on these devices.

\subsection{Motivation}

\begin{figure*}[ht!]
  \centering 
  \includegraphics[width=0.85\textwidth]{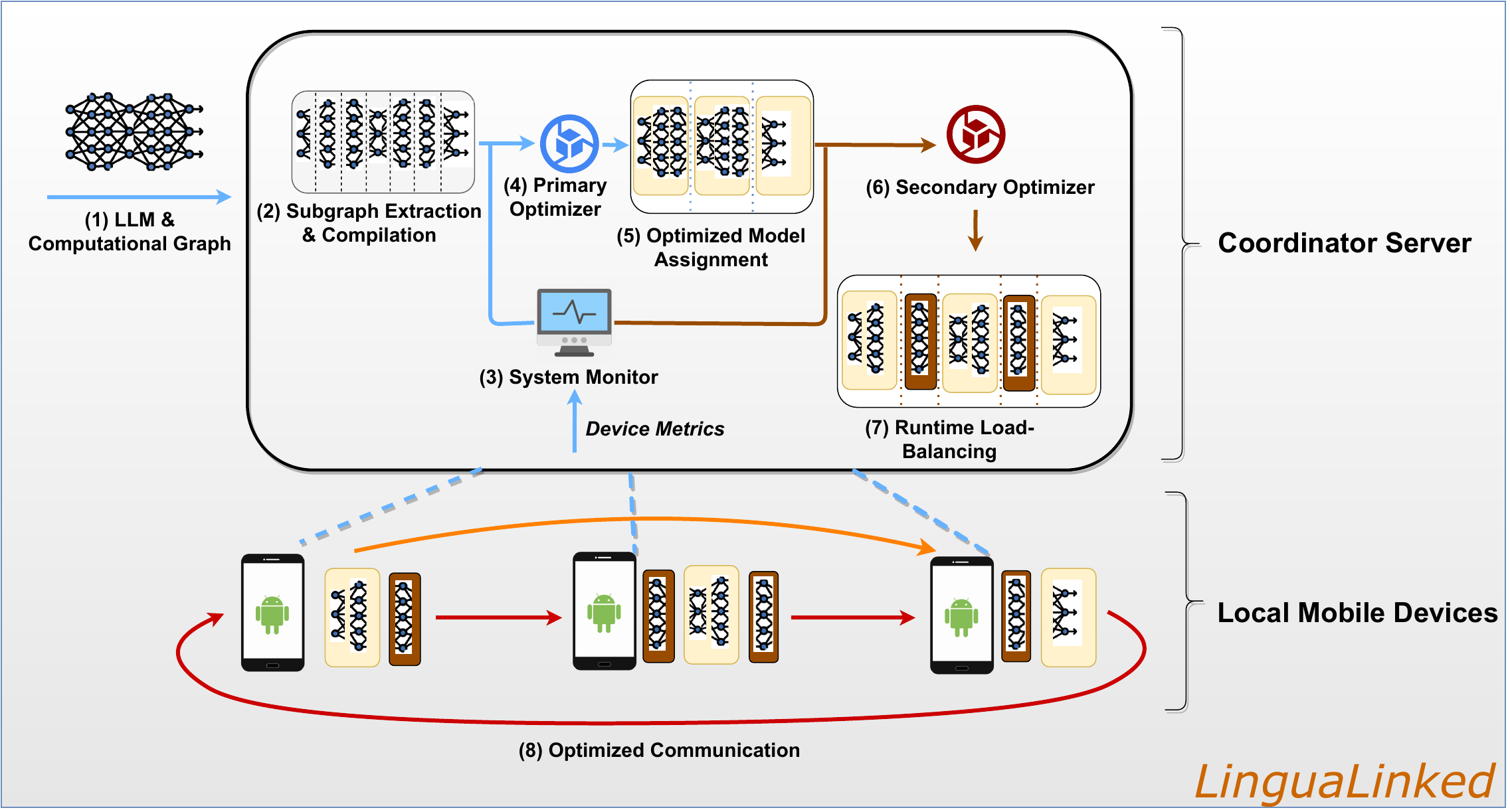} 
  \caption{Overview of \textit{LinguaLinked} System Design}
  \label{fig:overview}
\end{figure*}
\subsubsection{Distributed LLM Inference on Mobile Devices}
Given the memory and computational limitations of a single mobile device, a distributed inference approach emerges as a practical solution for LLMs. This methodology involves partitioning LLMs into smaller segments that align with the memory constraints of individual devices, a strategy implemented in \textit{LinguaLinked}. Each device is then responsible for processing a portion of the model, significantly reducing the memory burden on any single device and enabling a broader range of devices to participate in LLM inference tasks.

Due to the growing need for efficient and localized AI services, various distributed inference frameworks have been developed. DeepHome [\citealp{hu2019deephome}] implements data parallelism across diverse devices, while MODNN [\citealp{mao2017modnn}] partitions DNN models among multiple mobile units to speed up computation, though it does not consider the dynamic performance changes of devices in real-time. EdgeFlow [\citealp{hu2022distributed}] offers a framework suited for general DAG-structured deep learning models, predominantly in simulated environments. Recent work [\cite{xu2022distributed}] has also delved into load balancing algorithms for DNN inference across end devices and edge servers. However, despite these advancements, primarily in the context of vision models, there remains a notable gap in developing and implementing distributed inference strategies specifically for LLMs on mobile devices.

\subsubsection{Improved Data Privacy and Inference Latency}
The traditional model of centralized data centers requires transferring data over networks for processing. While this system has its strengths, it introduces two primary issues: increased privacy risks and latency in data processing. Centralized processing requires data to travel from the source to the data center and back, which can significantly delay response times, particularly in applications that require real-time processing. \textit{LinguaLinked} aims to address these challenges by decentralizing LLMs and deploying them directly on mobile devices. This local deployment means that data can be processed where it is generated, significantly reducing the need for data transmission over potentially untrusted networks. Consequently, this approach enhances data privacy and security, as the data remains within the user's local environment, reducing exposure to external threats.

\subsubsection{Scalability and Responsiveness in Distributed LLM Deployment}
In a distributed LLM environment, the variability in device capabilities and workloads can lead to inefficient distribution of computational tasks, potentially resulting in system bottlenecks and inconsistent performance. Load balancing is crucial in this context as it ensures that no single device is overwhelmed by inference tasks, thereby preventing performance degradation. This is particularly important in mobile computing, where devices have varying computational capacities and may not handle large workloads effectively. By dynamically redistributing tasks, load balancing helps maintain an optimal level of performance across all devices, ensuring that the system can scale effectively and respond to changing demands. This approach not only optimizes resource utilization but also contributes to the robustness and reliability of distributed LLM inference.

\section{System Design}
The \textit{LinguaLinked} system enables the distributed execution of LLMs on mobile devices. As shown in Figure~\ref{fig:overview}, the process begins with the LLM being loaded and transformed into a computational graph on a coordinator server (\circledsmall{1}). Subsequently, the server extracts the model subgraphs (\circledsmall{2}) and compiles the subgraphs into deployment-ready sub-modules. Once subgraph extraction and compilation are completed, the server analyzes mobile device metrics provided by the system monitor (\circledsmall{3}). Given the device performance metrics, a primary optimizer (\circledsmall{4}) provides an optimized model assignment strategy to allocate LLM sub-modules to mobile devices (\circledsmall{5}). A secondary optimizer (\circledsmall{6}) further refines the distribution of tasks by ensuring certain sub-modules are overlapped across devices to facilitate easy load balancing. The runtime load balancing strategy (\circledsmall{7}) reallocates tasks in response to runtime device performance metrics. Finally, the optimized communication strategy (\circledsmall{8}) ensures minimized data transmission between mobile devices during the inference process, thus maintaining the system's responsiveness. Note that throughout this process, data and intermediate activations remain on the mobile devices and are never sent to the server, thereby ensuring privacy.

\subsection{System Monitor}
\label{system_monitor}
The system monitor tracks metrics across and within individual devices. The monitor consists of two modules: the server module and the device module. The server module coordinates when and how the device monitors work on each of the devices. It periodically sends start and stop signals to devices in order to control whether the device will capture its metrics. It also receives data from the device modules and preprocesses the data to fit to the optimizer. The device module measures several performance indicators, including the bandwidth (\(\mathcal{B}\)) and latency (\(\mathcal{L}\)) of network communications between devices, available memory (\(\mathcal{M}_a\)), total memory (\(\mathcal{M}_t\)), and the processing speed in Floating Point Operations per Second (\(FLOP/s\)) on each device. The memory usage and inter-device latency are measured directly, while the bandwidth and \(FLOP/s\) are indirectly calculated using data transmission time and test model computation time.

To measure the bandwidth between two mobile devices, \textit{LinguaLinked} establishes a direct network connection between two devices, designated as device \(i\) (sender) and device \(j\) (receiver). After connection setup, device \(i\) transmits a predefined chunk of data to device \(j\). Upon receiving the data, the bandwidth \(\mathcal{B}_{i,j}\) is calculated on device \(j\) by \({\mathcal{B}_{i,j}} = \frac{{D}_{j}}{{T}_{j}}\)
where \({T}_{j}\) is the time from connection establishment to receiving all data in the size of \({D}_{j}\).

The measurement of \(FLOP/s\) is more complex since it involves deploying a test model across mobile devices. Each device receives a small test model and a tensor from the server. The device deploys it locally and monitors its execution. The device \(i\) will record the computation time \({T}_{i}\) of the model, given the number of FLOPs of the model as \(NumFlop\). The processing speed of device \(i\) can be calculated as \({FLOP/s_{i}} = \frac{NumFlop}{{T}_{i}}\).

\subsection{Optimized Model Assignment}

\subsubsection{Subgraph Extraction from LLMs}
\label{section:subgraph_extraction}
The initial phase in assigning LLM sub-modules to mobile devices involves transforming the LLMs into a computational graph that represents the entire neural network, with each operation within the network represented as a node. The main goal at this stage is to extract smaller subgraphs from the LLM's complete computational graph, each capable of functioning independently on separate devices. Note that the subgraph extraction process is conducted without consideration of the device status. 

Let \( G = (V, E) \) represent the computation graph of the LLM, where \( V \) is the set of nodes and \( E \) is the set of edges representing dependencies. We identify a node \( v \in V \) is a candidate node
\(c\) such that \(c \in C\) for subgraph partitioning if it satisfies the following conditions: 
\begin{equation}
\begin{aligned}
C = \left\{ v \in V \,\middle|\, \right. & \left. (\text{deg}^+(v) > 1) \,\,\,\wedge\, \right. \left. (\text{deg}^-(v) = 1) \,\right\}
\end{aligned}
\end{equation}
where \(\text{deg}^+(v)\) represents the out-degree of node \(v\). \(\text{deg}^-(v)\) denotes the in-degree of node \(v\), reflecting the number of edges entering \(v\) from other nodes. This suggests that \(v\) primarily processes input from a single node, typically corresponding to a single operation or layer. Nodes satisfying these conditions are primary candidates for LLM subgraph extraction because they often encapsulate distinct layers or components of the model, allowing for effective distribution across devices while preserving essential dependencies.

After pinpointing the candidate nodes \( C \), the graph is divided into a series of distinct subgraphs, denoted as \( S \). The subgraph extraction process is represented as follows:
\begin{equation}
\begin{aligned}
    S = \left\{ s \,\middle|\, \forall j, \right. & \left. s = \{ v \in V \mid l_{j} \leq v < l_{{j+1}}, \, s \subseteq G \right\}
\end{aligned}
\end{equation}
Each subgraph \( s \) in \( S \) represents a distinct segment of \( G \) and is guided by the boundaries established by layers of candidate nodes \( (l_{j}, l_{j+1}) \) from the set \( C \). The boundaries indicate that the subgraph \( s \) starts at the boundary of a layer of candidate notes \( l_j \) and extends to, but does not include, the boundary of the next layer of candidate nodes \( l_{j+1} \). This equation ensures that each subgraph \( s \) is an independent and coherent part of the model, in alignment with the overall structure of \( G \).

\subsubsection{Subgraph Dependency Search}
Nodes within a subgraph may depend on nodes in non-adjacent subgraphs. To manage dependencies, we introduce a subgraph dependency search algorithm. This algorithm generates two maps: the residual dependency map (\texttt{RDM}) and the sequential dependency map (\texttt{SDM}), to track dependencies among subgraphs in \( S \). The \texttt{SDM} records direct sequential dependencies between subgraphs, indicating dependencies on the immediately preceding subgraph. This is formalized as:
\begin{equation}
\begin{aligned}
\texttt{SDM}_{i,j} = \{ (u, v) \mid u \in s_{i}, \, v \in s_{j}, \, & \exists e \in E : e = (u, v), \\
& i = j - 1, \, s \in S \}
\end{aligned}
\label{eq:sdm}
\end{equation}
Here \(\texttt{SDM}_{i,j}\) represents a set of tuples \((u, v)\), where \(u\) is a node in subgraph \(s_{i}\) and \(v\) is a node in subgraph \(s_{j}\). The existence of an edge \(e\) in the edge set \(E\) connecting \(u\) to \(v\)
establishes the dependency of \(v\) on \(u\). The condition \(i = j - 1\) ensures that this dependency follows a sequential order, meaning the output of node \(u\) in subgraph \(s_{i}\) is required as an input for node \(v\) in the immediately preceding subgraph \(s_{j}\).

Additionally, the algorithm identifies residual dependencies by examining nodes not in the directly preceding subgraph but present in earlier ones. These are recorded in the \texttt{RDM}:
\begin{equation}
\begin{aligned}
\texttt{RDM}_{i,k} = \{ (u, v) \mid u \in s_{i}, \, v \in s_{k}, \, \exists e \in E : e = (u, v), \\
i < k, \, i \neq k-1, \, s \in S \}
\end{aligned}
\label{eq:rdm}
\end{equation}
where each element in \(\texttt{RDM}_{i,k}\) is a tuple \((u, v)\), where \(u\) is a node in subgraph \(s_{i}\) and \(v\) is a node in subgraph \(s_{k}\). Similarly, there exist an edge \(e\) in the edge set \(E\) connecting \(u\) to \(v\) establishes a dependency between these nodes. The conditions \(i < k\) and \(i \neq k-1\) ensure that this relationship is between nodes in non-adjacent subgraphs.

\subsubsection{Model Assignment Optimization} 
\label{section:optimized model partitioning}

The server compiles each subgraph \(s\) from the set \(S\) into a deployment-ready sub-module \(mod\) such that \(mod \in Mod\). This step transforms the computational graph into an executable format. Following this, the server undertakes a comprehensive profiling of each sub-module. This profiling encompasses the evaluation of three key metrics for each sub-module: \textbf{(1)} the number of floating-point operations (\(\textit{NumFlop}_{mod}\)), \textbf{(2)} the memory requirement (\(\mathcal{M}_{mod}\)), and \textbf{(3)} the size of data output quantified in bytes (\(\mathcal{O}_{mod}\)). With this data, the system monitor, as detailed in Section~\ref{system_monitor}, begins tracking the mobile device's performance metrics and relays this information back to the server for the start of model partitioning optimization. 

The primary optimization target of our system is to find the optimized model assignment strategy for each device to minimize the total inference time. We formulate our optimization target as a linear optimization problem and use a linear programming solver (LP solver) to optimize the model partitioning process. We view the total inference time \(T_{total}\) as the sum of local computation time \(T_{compute}\) and data transmission time \(T_{data}\). Formally, the optimization process is represented as follows:
\begin{equation}\label{LP1}
~ \text{minimize} ~ (T_{compute}  ~ +  ~ T_{data})
\end{equation}

\text{~~~where}~ \[T_{compute}= \sum_{i=0}^{\text{m}} \sum_{j=0}^{\text{n}} \mathcal{E}_{i,j} \cdot x_{i,j}\]

\[T_{data} = \sum_{i=0}^{\text{m}} \sum_{\substack{j=0\\ i \neq j}}^{\text{m}} \left( \mathcal{L}_{i,j} + \frac{\mathcal{O}^{d2d}_{i,j}}{\mathcal{B}_{i,j}} \right) \]

\[\mathcal{E}_{i,j} = \frac{NumFlop_{mod_{j}}}{FLOP/s_{i}}\]

\[\mathcal{O}^{d2d} =  X \cdot \mathcal{O}^{m2m} \cdot X^T \]

We define the set of devices \( \mathcal{D} = \{d_0, d_1, \ldots, d_{m}\} \) to represent the \( m \) available devices and the set of sub-modules \( Mod = \{mod_0, mod_1, \ldots, mod_{n}\} \) to represent the \( n \) sub-modules resulting from the initial model partitioning. The optimized model assignment strategy \( X \in \{0, 1\}^{m \times n} \) is a Boolean variable matrix indicating the assignment of sub-modules to devices, where the element \( x_{i,j} \) signifies whether sub-module \( mod_j \) is assigned to device \( d_i \).

For measuring the \(T_{compute}\) across all sub-modules and devices, we divide the number of FLOPs of the sub-module \(mod_n\) and the processing speed in FLOP/s of the device \(d_m\). Here, \(\mathcal{E} \subseteq \mathbb{R}^{m \times n}\) represent the sub-module local computation time matrix, where each element \(\mathcal{E}_{i,j}\) denotes the time taken by device \(d_i\) to execute sub-module \(mod_j\). For measuring the \(T_{data}\), we focus on the communication latency time and the data transmission overhead between devices. \(\mathcal{L} \subseteq \mathbb{R}^{m \times m}\) and \(\mathcal{B} \subseteq \mathbb{R}^{m \times m}\) are matrices representing the communication latency and bandwidths between the \(m\) devices. The term \(\mathcal{O}^{d2d} \subseteq \mathbb{R}^{m \times m}\) refers to the output size associated with the devices, which is inferred from \(\mathcal{O}^{m2m} \subseteq \mathbb{R}^{n \times n}\), indicating the output size from one sub-module to another.

The LP has the following memory constraints:
\[ \forall i \in \{0,\ldots, m\} \sum_{j=1}^{\text{n}} \mathcal{M}_{mod_{j}} \cdot x_{i,j} \leq \beta \cdot \mathcal{M}_{a_{i}} \]
\( \mathcal{M}_{mod} \in \mathbb{R}^n \) represents the memory requirement of each sub-module. The coefficient \( \beta \) specifies the maximum proportion of memory that each device can allocate for the inference task. This constraint ensures that the sub-modules assigned to any device \( d_i \) do not exceed \( \beta \) times the available memory \( \mathcal{M}_{a_i} \) of that device.

\begin{figure}
  \centering 
    \includegraphics[width=\linewidth]{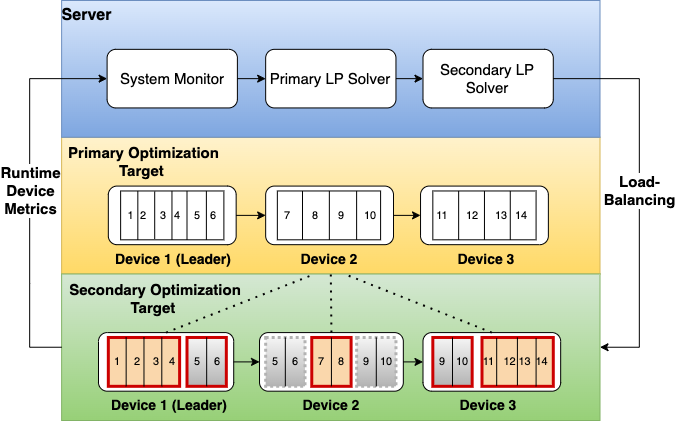} 
  \caption{Runtime Load Balancing}
  \label{fig:lb}
\end{figure}
\subsection{Runtime Load Balancing}
\label{section:runtime_load_balancing}
\subsubsection{Load Balancing Optimization}
The load balancing mechanism is the secondary linear optimization problem, which refines the primary optimization target \( X \) in \S~\ref{section:optimized model partitioning}.

Our load balancing optimization employs a strategy that involves overlapping sub-module deployment across devices to improve system performance. In this strategy, we classify sub-modules on each device into two types: \textbf{movable} and \textbf{unmovable}. The movable sub-modules are designed for dynamic loading or offloading during load balancing across adjacent devices, while the unmovable sub-modules are fixed and remain permanently active on their assigned devices. In Figure~\ref{fig:lb}, the numbers indicate the index of sub-modules that are assigned to each device. Sub-modules highlighted with a solid border are the ones that are currently active in memory. The sub-module indices that appear on multiple devices are the movable sub-modules, overlapped and deployed across two adjacent devices. Inactive submodules are not loaded into memory but are stored on secondary storage instead. The load balancer dynamically manages their memory allocation, triggering their loading into memory as needed.

Integral to our load balancing approach is the linear programming formulation, which minimizes data transmission between devices and balances the load. It achieves this by computing the optimal output and input sizes for each sub-module during the model partitioning phase. Consequently, this optimization results in more efficient data management and improves the overall robustness of the system, especially during intensive inference tasks.

Given the primary optimization target \( X \in \{0, 1\}^{m \times n} \), we introduce two new boolean variables, \( \hat{X}_l \in \{0, 1\}^{m \times n}\) and \( \hat{X}_r \in \{0, 1\}^{m \times n}\), as the secondary optimization targets. They are defined to represent the potential for left and right overlaps of sub-modules on devices. The variables \( \hat{X}_r \) and \( \hat{X}_l \) are initialized based on \( X \) to reflect this potential overlap, with \( \hat{X}_r \) denoting the sub-modules to the right of the last allocated sub-module and \( \hat{X}_l \) denoting those to the left of the first allocated sub-module on each device.

To optimize the overlap, we solve a secondary LP. The optimization seeks to maximize the memory utilization within the constraints of device memory, thereby improving load balancing across devices. This is formalized as follows:
\begin{align}\label{LP2}
& ~ \text{minimize} ~(\mathcal{M}_{left}  ~ + ~ \mathcal{M}_{right}) \\
& \text{i.e., minimize} ~ \sum_{j=0}^{\text{n}} \mathcal{M}_{mod_{j}} \cdot \lbrack(\hat{x}_{l_{i,j}} \lor x_{i,j})  + (\hat{x}_{r_{i,j}} \lor x_{i,j})\rbrack
\end{align}
subject to the memory constraints for each device:
\begin{equation*}
\forall i \in \{0,\ldots, m\}, \sum_{j=0}^{\text{n}} \mathcal{M}_{mod_{j}} \cdot (\hat{x}_{l_{i,j}} \lor x_{i,j} \lor \hat{x}_{r_{i,j}}) \leq \beta \cdot \mathcal{M}_{a_{i}}
\end{equation*}
Each device may have left overlapping modules that can be moved to the preceding device and right overlapping modules that can be reloaded on the succeeding device. The variable \( x_{i,j} \) is an element of the sub-module allocation matrix \( X \), indicating whether the $ j^{th} $ sub-module is allocated to the $i^{th}$ device. The binary variables \( \hat{x}_{l_{i,j}} \) and \( \hat{x}_{r_{i,j}} \) indicate whether the $ j^{th} $ sub-module overlaps on the left or right side, respectively, on the $i^{th}$ device. The constraints ensure that the memory is not overloaded.

\subsubsection{Model Deployment with Load Balancing}

LinguaLinked decide whether to trigger load balancing based on the runtime device metrics collected by the system monitor. When the load is imbalanced, the solution to the initial LP (Eq.~\ref{LP1}) is used alongside the secondary optimization targets \( \hat{X}_r \) and \( \hat{X}_l \) to derive the new assignment. The unmovable modules have a fixed assignment in the secondary LP. The movable/overlapping module locations are determined by the solution to the load balancing LP (Eq.~\ref{LP2}).

After finalizing the new allocation strategy, the load balancer directs each device to modify its load given the updated strategy. When transitioning to a new allocation, the computational activities are paused at the individual device level rather than across the entire pipeline, thereby minimizing disruptions. The transition to the new pipeline occurs in a sequential manner across devices while other devices in the pipeline continue their operations uninterrupted. Once the first device (Leader) finishes its ongoing tasks under the previous allocation strategy, it transitions to the new strategy. Subsequent devices in the chain follow suit, adopting the new tasks only after completing their current processing. This sequential transition from one allocation strategy to another ensures an efficient re-optimization of the inference process, minimizing disruption and maintaining continuous operation throughout the pipeline.

\subsection{Optimized Communication}
\label{network_communication_section}
\subsubsection{Decentralized Device Communication}
\begin{figure}
  \centering 
    \includegraphics[width=\linewidth]{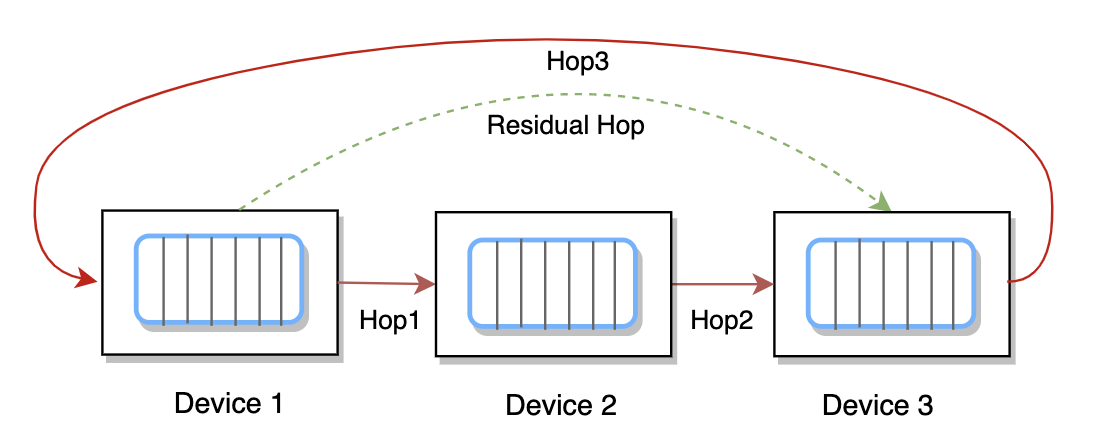} 
  \caption{System Design For Device Communication.}
  \label{fig:res}
\end{figure}

In our system, devices communicate in a decentralized manner, forming a ring structure as depicted in Figure ~\ref{fig:res}. This process initiates from the leading device, referred to as the Leader, which transmits its output to the next device in the sequence. Once the final device in the ring completes its computation, it sends the results back to the Leader. Figure ~\ref{fig:res} illustrates this data flow with a solid red line, showing the progression from the lead device to the final one, and then back to the start.

The communication among these devices is efficiently managed using a message queue that implements the \texttt{ROUTER}-\texttt{DEALER} pattern, known for its scalability in communication frameworks. In this setup, the \texttt{ROUTER} socket is responsible for distributing messages to various \texttt{DEALER} sockets. Conversely, the \texttt{DEALER} sockets send messages back to the \texttt{ROUTER} asynchronously. This pattern is ideal for scenarios where a central node (the \texttt{ROUTER}) needs to manage communication with multiple worker nodes (\texttt{DEALERs}), thereby ensuring balanced load distribution and efficient task allocation. In our system, each device alternates between being a \texttt{ROUTER} (sending data) and a \texttt{DEALER} (receiving data), depending on its position in the sequence.

The process within each device is two-fold: upon receiving the output from the previous device, the device (now acting as the receiver) begins its inference computation. After completing this task, the device (switching to the sender role) forwards the computed results to the next device in the ring. This systematic flow ensures continuous and organized data processing throughout the network.
\subsubsection{Multi-Threaded Inference}
To improve the system efficiency and achieve better resource utilization, we implement multi-threaded inference, where each thread independently handles a task, facilitating parallel processing. This configuration enables the lead device, and consequently, each thread, to immediately move on to the next task after completing the current one. During the initial setup, we prioritize the initialization of an appropriately sized thread pool. Given the non-threadsafe nature of message queue sockets, special attention is required to maintain thread safety in a multi-threaded context. To circumvent potential issues, we avoid sharing sockets across threads. However, considering that locking sockets might hinder parallelism in data transmission, we generally prefer the use of multiple sockets and ports. This approach not only avoids performance bottlenecks but also aligns with our goal of minimizing communication latency.

Multi-threaded inference plays a crucial role in maximizing CPU efficiency. It accommodates varying batch sizes and enables the immediate processing of samples upon their receipt. We have explored different strategies for batch processing in multi-threaded environments, such as dividing a large batch into smaller mini-batches for individual threads or allocating different batch sizes to various threads. This flexibility allows for dynamic adjustment of batch sizes, further optimizing our system's performance.

\subsubsection{Sequential \& Residual Communication}
\label{seq_res_communication}
In sequential communication, each device sends its output to the subsequent device and can only receive data from the previous device, see the solid line in Figure~\ref{fig:res}. This creates a circuit communication where data flows strictly from one device to the next in sequence. Once a device completes its inference and sends all outputs forward, it can begin processing new samples. However, there is a limitation to this communication method.  In this case, the residual data is piggybacked with the sequential data and passed through by the recipient devices until the target is reached. It incurs the latency costs associated with receiving and transmitting this unused data. Particularly, if the data is only utilized by a device much later in the sequence, all preceding devices unnecessarily carry and relay this data from start to the target device.

To address this, we have incorporated a residual communication strategy. The green dashed line in Figure~\ref{fig:res} illustrates the residual connections. The \texttt{SDM} (Eq.~\ref{eq:sdm}) and \texttt{RDM} (Eq.~\ref{eq:rdm}) provide a comprehensive understanding of both sending and receiving dependencies across sub-modules. Utilizing these maps, we generate a device dependency map that guides each device in accumulating outputs from its sub-modules and efficiently routing them to the correct receiving devices.

A sub-module may produce only sequential output, with no residual output. On the other hand, multiple sub-modules on one device might have residual outputs to the same target device. In such scenarios, these residual outputs are aggregated to enhance the efficiency of data transfer. This aggregation is facilitated through shared communication sockets for transmitting the combined residual outputs of different sub-modules to the same target device, thereby reducing the complexity and overhead associated with maintaining multiple channels and improving network resource utilization.

To ensure timely and efficient data delivery, the transmission order is pre-arranged. The nearest device in the network topology receives its residual data first, minimizing latency. This process is managed through distinct threads for both sequantial and residual data transmission and reception, ensuring that residual data handling does not interfere with the primary computational tasks of each device. After the successful reception of all residual data, the threads are terminated to free up system resources. However, the communication sockets remain active. They are established prior to the start of the entire inference process, allowing for continuous and ready data transfer in subsequent cycles.

\section{Evaluation}
\begin{table}[ht]
    \centering
    \caption{Test Hardware Platforms in Evaluation.}
    \label{device-info}
    \small
    \begin{tabular}{lccl}
    \hline
        & Pixel 7 pro  & CUBOT X30\\ \hline
    SoC & Google Tensor G2      & Mediatek MT6771 Helio P60  \\ 
    CPU & Cortex-X1/A78/A55      & Cortex-A73/A53 \\
    RAM & 12GB         & 8GB    \\
    OS  & Android 13   & Android 10 \\\hline
    \end{tabular}
    
\end{table}

\subsection{Implementation and Methodology}
\textbf{LinguaLinked Prototype.} We build LinguaLinked atop PyTorch \cite{paszke2019pytorch}. We use the \texttt{torch.fx} library \cite{reed2022torchfx} to extract computational subgraphs and compile them into executable Pytorch sub-modules. To profile these sub-modules, we employ Deepspeed \cite{rasley2020deepspeed}, which provides us with detailed information like the FLOPs and memory footprints. Following this, the sub-modules are converted into the ONNX \cite{bai2019} format, making them suitable for mobile device deployment. To facilitate this further, we apply dynamic quantization using the ONNXRuntime \cite{ONNXRuntime} library. We specifically use \texttt{int8} precision quantization, as ONNXRuntime only supports \texttt{int8} models for mobile platforms.

For the optimization of model assignment and runtime load balancing, we use Gurobipy \cite{gurobi}, a Mixed-Integer Linear Programming (MILP) solver. With these optimizations, the LLM sub-modules are ready for deployment on mobile devices. In the deployment phase, we have developed an Android application that leverages the ONNXRuntime C++ API to deploy the ONNX formatted sub-modules. To ensure efficient communication and distributed inference on mobile devices, we integrate ZeroMQ \cite{ZeroMQ24:online}, a high-performance asynchronous messaging library. Specifically, we use the ROUTER-DEALER socket pattern in ZeroMQ, which offers an asynchronous, non-blocking alternative to the standard REQ-REP sockets. This design choice is crucial for the effective interaction among various devices in our setup.\\
\textbf{Hardware.} In our evaluation, we employ 4 mobile devices: 3 units of the high-end Google Pixel 7 Pro and 1 unit of the low-end CUBOT X30. The specific hardware configurations of these devices are detailed in Table~\ref{device-info}. Our system performance evaluation is conducted exclusively on the CPU of these devices. The current implementation of our system aligns with the capabilities of ONNXRuntime, which currently supports CPU-based operators for LLMs. This alignment is strategic, considering that ONNXRuntime’s GPU acceleration for LLM tasks is still under development. We have designed the LinguaLinked system with flexibility and scalability in mind, ensuring that once ONNXRuntime extends its support to include GPU acceleration, our system will seamlessly integrate into mobile GPU environments without requiring significant modifications. Consequently, we focus on CPU performance to show the system's efficacy under these constraints and to ensure a consistent evaluation environment across different device models.\\
\textbf{Evaluation Tasks.} We evaluate our system using two main tasks: text generation and text classification. For the text generation task, we utilize the Wikitext-2 dataset \cite{merity2016pointer}. Due to the computational limitations of mobile devices, we randomly select 100 samples from this dataset. For each sample, we set a context length of 20 tokens and generate 50 tokens.

As a secondary evaluation task, we focus on text classification using the IMDB sentiment classification dataset \cite{maas-EtAl:2011:ACL-HLT2011}. Here, we also randomly sample 100 instances from the dataset. In this task, each sample's entire context is used for classification. These tasks have been chosen and designed to comprehensively evaluate our system's capabilities within the constraints of mobile device computing power.\\
\textbf{Test Models.} For our system evaluation, we utilize the BLOOM series of LLMs \cite{workshop2023bloom}, developed by Hugging Face for a range of Natural Language Processing tasks. The models in this series are distinguished by their parameter sizes, and we use the models with 3 billion, 1.7 billion, and 1.1 billion parameters, respectively. We refer to these models as BLOOM 3b, BLOOM 1.7b, and BLOOM 1.1b. Each model is assessed in two precision formats: the full precision and the \texttt{int8} precision form.\\
\textbf{Baseline for Comparison.} Given the absence of existing research specifically focused on on-device distributed inference for LLMs, we have taken the initiative to design our own baseline experiments in order to assess our system's performance effectively. This baseline involves a straightforward assignment of sub-modules across the available mobile devices: given the total number of sub-modules \(m\) and the total number of mobile devices \(n\), each device is assigned an equal share of \(m/n\) sub-modules. This experiment is designed to be independent of individual device hardware specifications and network status, providing a uniform distribution of computational load. To show the effectiveness of our system, we compare the throughput achieved in this baseline scenario against the results obtained using our optimized model assignment strategy. Additionally, we compare it with the outcomes of the runtime load balancing strategy.

\begin{figure*}[ht!]
  \centering
  \begin{subfigure}[b]{0.35\textwidth}
    \includegraphics[width=\linewidth]{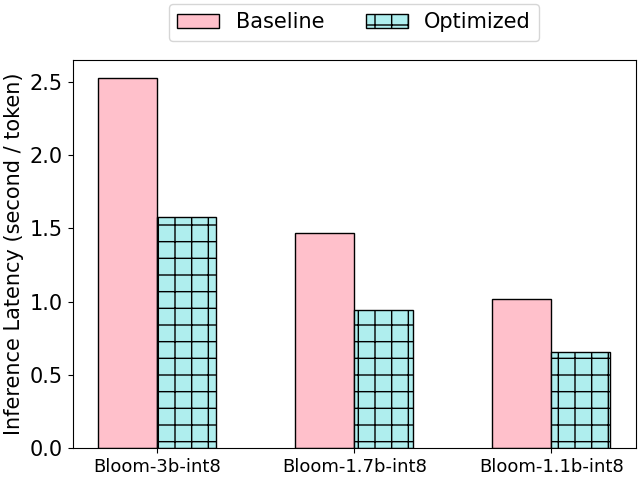}
    \caption{\texttt{int8} Inference Throughput for Text Generation (2 Pixel 7 pro, 1 Cubot X30).}
    \label{fig:quant-opt}
  \end{subfigure}
  \hspace{1cm}
  \begin{subfigure}[b]{0.35\textwidth}
    \includegraphics[width=\linewidth]{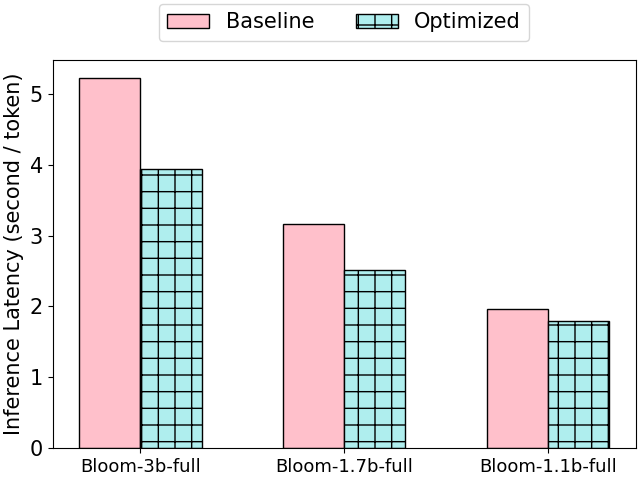}
    \caption{Full Precision Inference Throughput for Text Generation (3 Pixel 7 pro, 1 Cubot X30).}
    \label{fig:full-opt}
  \end{subfigure}
  \caption{Comparison of Baseline and Optimized Strategies for Model Assignment in Heterogeneous Devices.}
  \label{fig:mainfigure}
\end{figure*}

\begin{figure*}[h]
  \centering 
  \includegraphics[width=\textwidth]{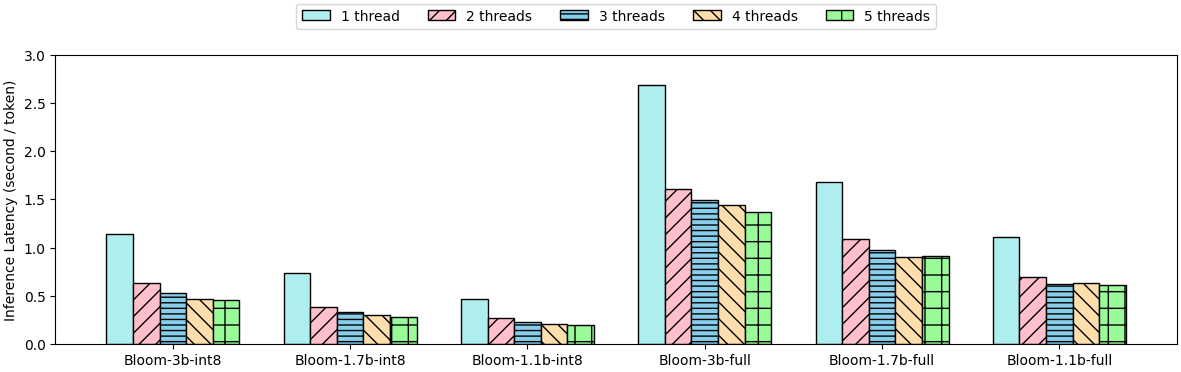} 
  \caption{Multi-threading Throughput for Text Generation on 3 Google Pixel 7 pro.}
  \label{fig:multi-thread-generation}
\end{figure*}

\subsection{Performance}
\textbf{Optimized Model Assignment Performance.} We compare the efficiency of optimized model assignment strategies by analyzing the inference throughput across baseline and optimized assignments in a heterogeneous device environment. For the \texttt{int8} quantized models, the inference throughput is assessed using 3 mobile devices: 2 Google Pixel 7 Pro smartphones and 1 Cubot X30, whereas for the full-precision models, the evaluation is conducted on 4 mobile devices: 3 Google Pixel 7 Pro smartphones and 1 Cubot X30.

For the quantized models, the optimized assignments lead to significant improvements in throughput, as shown in Figure~\ref{fig:quant-opt}. With the BLOOM 3b quantized model, LinguaLinked offers a 1.61\(\times\) increase in inference throughput over the baseline. The BLOOM 1.7b and 1.1b quantized models show similar levels of improvement, with 1.55\(\times\) and 1.56\(\times\) increase, respectively. The full-precision models also benefit from the optimized model assignments in LinguaLinked, as shown in Figure~\ref{fig:full-opt}. The BLOOM 3b full-precision model sees a \(1.32\times\) improvement in inference throughput. The BLOOM 1.7b and 1.1b models also follow this trend, with the former achieving a \(1.25\times\) increase and the latter a \(1.11\times\) increase.

We observe that as the size of the model increases, the improvement in inference throughput from optimized model assignment becomes more significant. The BLOOM 3b model, the largest evaluated model, demonstrates the highest throughput increase, suggesting that optimization strategies become increasingly effective for larger models, which have higher computational demands. This pattern indicates that optimized model assignment is particularly beneficial in enhancing the performance of larger models in distributed inference tasks within heterogeneous computing environments.

\begin{figure}[ht]
  \centering 
  \includegraphics[width=\linewidth, height=5.5cm]{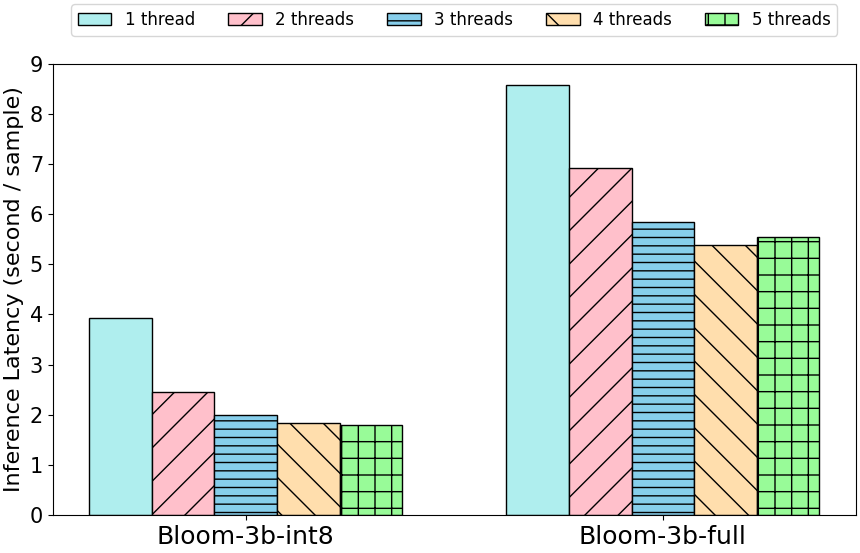} 
  \caption{Multi-threading Throughput for Text Classification on 3 Google Pixel 7 pro.}
  \label{fig:multi-thread-classification}
\end{figure}

These results underscore the efficacy of optimized model assignment strategies for distributed inference, especially when devices are heterogeneous. While the quantized models significantly benefit from optimization, full-precision models also show notable improvements, indicating that careful resource allocation and computational load balancing are advantageous across different model sizes and precision.\\
\textbf{Multi-threaded Inference Performance.} We evaluate the impact of multi-threading on inference throughput using both quantized (\texttt{int8}) and full-precision variants of BLOOM series models, with a focus on text classification tasks for the BLOOM 3b model and text generation tasks across the BLOOM 3b, 1.7b, and 1.1b models. We conduct the experiments on three Google Pixel 7 Pro devices. Performance improvements for text generation and text classification are presented in Figure~\ref{fig:multi-thread-generation} and Figure~\ref{fig:multi-thread-classification}, respectively.\\
For text generation tasks, the speed-up of quantized models is particularly significant. The BLOOM 3b quantized model demonstrates a remarkable increase in inference throughput. When comparing the single-threaded setup to a dual-threaded one, the average compute time per token improves by approximately 1.81\(\times\). As we increase the thread count to five, the inference throughput increases to 2.52\(\times\) faster than the single-thread baseline. This substantial gain underscores the efficiency of multi-threading in distributed quantized model inference. Similarly, the BLOOM 1.7b and 1.1b quantized models exhibit significant speed-ups. The BLOOM 1.7b model's inference throughput nearly doubles (1.90\(\times\) faster) with two threads, and with five threads, it becomes 2.65\(\times\) faster than the single-threaded approach. The BLOOM 1.1b model sees a speed-up of approximately 1.73\(\times\) with two threads, and by extending to five threads, the speed-up is 2.3\(\times\) that of the single-threaded performance. 

In contrast, the full-precision models exhibit a less drastic but still notable improvement with multi-threading. The BLOOM 3b full-precision model's average compute time per token decreases by 1.67\(\times\) when moving from one to two threads. Increasing the thread count to five yields a speed-up of 1.97\(\times\) compared to the single-threaded scenario. The BLOOM 1.7b and 1.1b full-precision models follow a similar pattern, with speed-ups of 1.54\(\times\) and 1.58\(\times\) respectively with two threads. With five threads, these models achieve a speed-up of 1.83\(\times\) and 1.79\(\times\), respectively, over their single-threaded counterparts.

Similar to the text generation task, for the text classification task, quantized models show significant efficiency gains. The BLOOM 3b quantized model exhibits speed-up in throughput. Shifting from a single-threaded to a dual-threaded setup, the average compute time per sample is reduced by 1.60\(\times\), and expanding to five threads yields 2.18\(\times\) acceleration compared to the single-thread baseline. The full-precision model benefits from multi-threading as well. The BLOOM 3b full-precision model sees a decrease in average compute time per sample by 1.24\(\times\) with two threads, and a 1.55\(\times\) increase in throughput with five threads relative to a single-threaded setup.

It is evident that the performance improvement from multi-threading is not linear; it exhibits a trend of diminishing returns as more threads are added. The initial shift from one to two threads provides the most significant performance boost. Beyond this point, while additional threads continue to improve inference speed, the rate of speed-up incrementally decreases. Thus, while multi-threading is an effective strategy for improving throughput, the optimal number of threads must be determined empirically to ensure the best use of computational resources without incurring unnecessary overhead.

\begin{figure}
  \centering 
    \includegraphics[width=0.9\linewidth, height=5.5cm]{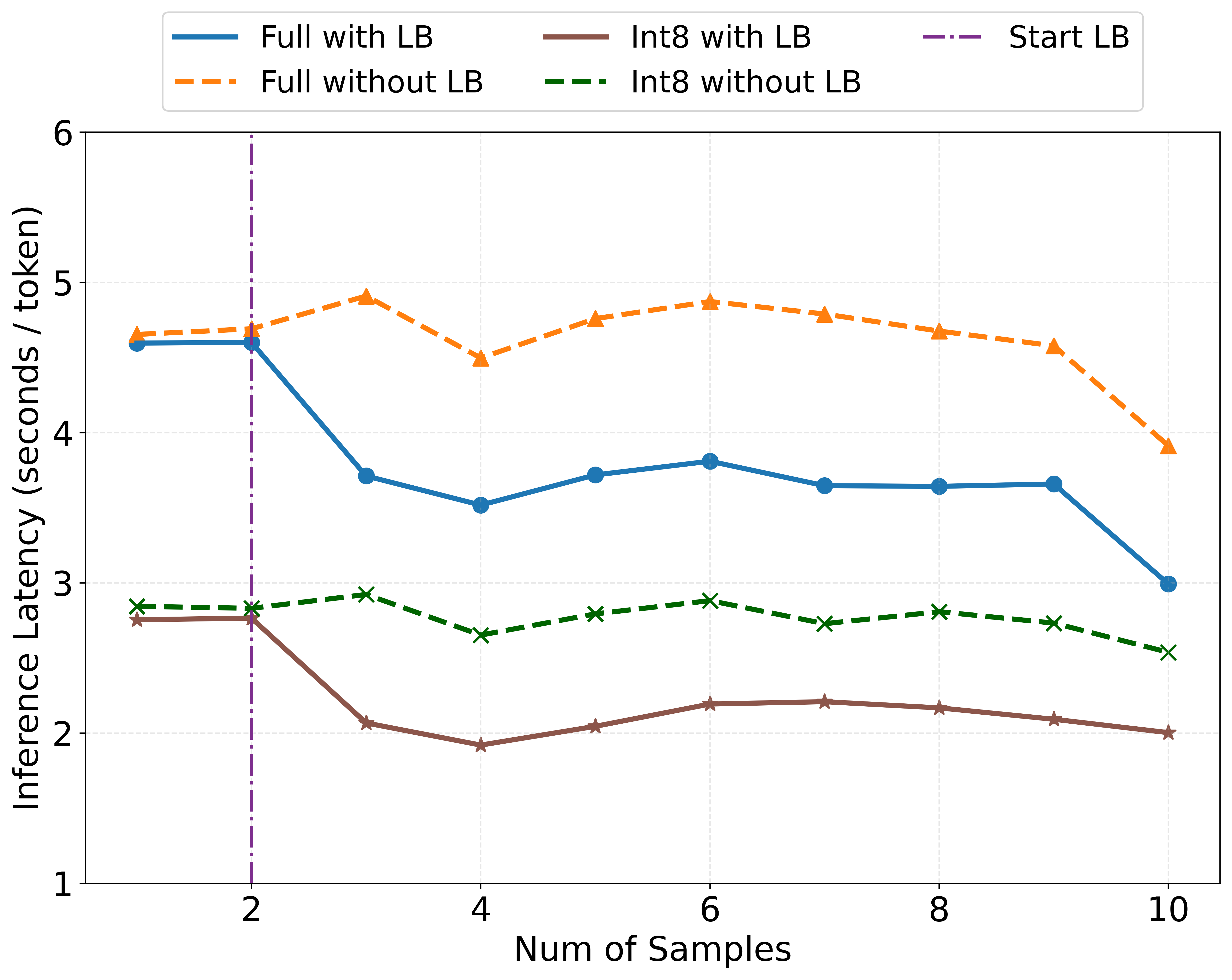} 
  \caption{Load Balancer Launched at Runtime.}
  \label{fig:runtime_lb}
\end{figure}

\textbf{Micro-Benchmarking the Runtime Load Balancer.}
We assess the impact of runtime load balancing on both the BLOOM 1.7b full precision and \texttt{int8} quantized models. We use three devices: two high-end phones and one low-end phone. In this setting, the initial model assignment assigns the first and last partitions to the high-end phones and the middle partition to the low-end phones. The load balancer is disabled during the beginning of the experiment. In this instance, the low-end phone in the middle is overloaded. We enable the load balancer after 2 samples are processed and continue processing 10 samples and measure the time taken per token.

As shown in Figure ~\ref{fig:runtime_lb}, activating load balancing starting from the second sample leads to a noticeable reduction in inference latency. Prior to load balancing, the inference latency for the BLOOM 1.7b full precision model is approximately 4.624 seconds per token for both scenarios, with and without load balancing. When we enable the load balancing module after the second sample, the latency decreased to around 3.587 seconds per token, resulting in an average reduction of approximately 1.036 seconds per token. We observe that a total of 8 sub-modules were relocated when the load balancing was enabled. The overhead associated with load balancing for the full precision model, which involves releasing these overlapping segments on the overloaded device, amounts to 0.03 seconds. However, the process of reloading these sessions onto other devices takes significantly longer, approximately 2.449 seconds in total, making it a substantial overhead in the context of load balancing. 

Similarly, for the BLOOM 1.7b quantized model, the inference time is approximately 2.756 seconds without load balancing. However, when we activate the load balancer at the second sample, it reduces the inference time by approximately 0.670 seconds per token on average, resulting in an inference time of 2.087 seconds per token. The improvement in inference latency demonstrates the effectiveness of load balancing in making runtime tasks more efficient than evenly partitioning the workload among devices with varying capabilities. When dealing with the 8 quantized sub-modules, the load balancing overhead for the quantized model consists of 0.018 seconds for releasing and 0.234 seconds for loading. The reloading time for the quantized model is significantly shorter than that for the full precision model, given the memory footprint of 8 full precision modules being four times larger than that of 8 quantized submodules. In this context, reloading sessions is the primary overhead in the process for load balancing in comparison with releasing session.

For the same set of samples, the inference time for the quantized model is slightly more stable than that of the unquantized model in the generation tasks. Comparing the performance of the BLOOM 1.7b full precision and quantized models, the average reduction in time for the full precision model was larger than that of the quantized model. This is reasonable because the full precision model is approximately $4\times$ larger than the quantized model, resulting in larger overlapped modules and more computational time being offloaded from low-level devices to high-level devices. However, the acceleration rate for the quantized model is approximately 1.32$\times$, while for the full precision model, it is around 1.29$\times$, which is quite close. On average, there was a 30\% improvement in acceleration when we initiated load balancing during runtime.

\subsection{Sequential and Residual Communication}
To illustrate the benefits of residual communication over sequential communication, we measure the time for each communication instance. Given the small size of residual data and the ample bandwidth of the devices, achieving highly synchronized measurements for hop time is essential. To accomplish this, we leverage the Network Temporal Protocol (NTP) to synchronize the starting time of each device down to the millisecond level and rely on the system clock to measure the actual runtime with nanosecond precision.

\begin{table}[ht]
    \centering
    \caption{Residual and Sequential Communication Performance}
    \begin{tabular}{|l|cccc|c|}
    \hline
        & Hop1  & Hop2 & Hop3 & Total     & Res Hop          \\ \hline
    Seq & 0.2489s  & 0.2580s & 0.0770s  & 0.5839s & ------ \\ 
    Res & 0.2347s & 0.2463s & 0.0779s  & 0.5589s  & 0.0111s     \\\hline
    \end{tabular}
    \label{seqvsres}
\end{table}
To demonstrate the differences between sequential and residual communication, we select three low-end smartphones with lower bandwidth capabilities and employ the quantized BLOOM 3b model. We report the average delay from ten experiments, with each query generating a response with a length of 100 tokens. In Table ~\ref{seqvsres}, we can observe several key metrics. Hop1 is the sequential transmission time between device 1 and device 2, and Hop2 between device 2 and device 3. Additionally, Hop3 encompasses the time required for sending logits back from device 3 to device 1, seen in figure ~\ref{fig:res}. Residual Hop measures the transmission of residual data directly from device 1 to device 3.

The sequential communication has higher delay for two reasons: (i) multiple hops instead of the direct residual hop, and (ii) the residual data is piggybacked along with activations which are typically much larger in size and need to wait for inference to complete in relaying devices. Note that as the number of residual connections and the size of the model increase, the saved time will likewise increase. The average data transmission size measured for Hop1 and Hop2 is approximately 1.5 MB (including activations), while the size of the residual data for Residual Hop is around 15 KB. Moreover, Hop1, Hop2, and Hop3 must be processed sequentially while Residual Hop operates in parallel with the other Hops without waiting for the previous computing.

\section{Discussion}
Our results demonstrate promising advancements in distributed LLM inference on mobile devices but also underscore several challenges. Key among these are the overheads from load balancing and constraints of current hardware and software frameworks. As tools like ONNXRuntime evolve to support GPU acceleration, we expect significant enhancements in \textit{LinguaLinked's} performance. Furthermore, exploring advanced quantization techniques and communication mechanisms could lead to more efficient distributed inference systems.

A major direction for expanding \textit{LinguaLinked} involves adapting it for distributed fine-tuning on mobile devices, allowing model customization based on user interactions and local data, paving the way for personalized AI applications while preserving data privacy. We also envision extending \textit{LinguaLinked} to handle multi-modality models, enhancing its applicability in diverse real-world scenarios.

To further improve \textit{LinguaLinked}, we envision more advanced model computational graph partitioning strategies involving further optimizations on task divisions better aligned with device capabilities. Moreover, integrating advanced load balancing algorithms that account for not only computational capabilities but also battery life and user engagement patterns will ensure a holistic approach to distributed computing on mobile platforms.

Finally, a critical focus for future iterations of \textit{LinguaLinked} is energy efficiency. We find that the continuous intensive inference tasks, especially with full-precision models, significantly drain battery life and cause overheating, leading to performance degradation. To address this, we aim to incorporate energy-efficient computing strategies that balance computational demands with energy consumption and thermal management. This could include adaptive algorithms to modulate computational load based on the device's energy state, and hardware-specific optimizations leveraging low-power processing cores for specific tasks. 

\section{Conclusion}
In this work, we introduce \textit{LinguaLinked}, a system for decentralized LLM inference on mobile devices. To the best of our knowledge, \textit{LinguaLinked} is the first work that exploits deploying LLM distributively on mobile devices. \textit{LinguaLinked} implemented optimized model assignment strategy, network communication and runtime load balancing mechanism to accelerate the distributed LLM inference on mobile devices. This approach tackles the complexities of deploying both full precision and quantized LLMs of various sizes within mobile computing environments.

\bibliographystyle{ACM-Reference-Format}
\bibliography{sample-base}


\begin{thebibliography}{70}

%%% the Web address.

\ifx \showCODEN    \undefined \def \showCODEN     #1{\unskip}     \fi
\ifx \showDOI      \undefined \def \showDOI       #1{#1}\fi
\ifx \showISBNx    \undefined \def \showISBNx     #1{\unskip}     \fi
\ifx \showISBNxiii \undefined \def \showISBNxiii  #1{\unskip}     \fi
\ifx \showISSN     \undefined \def \showISSN      #1{\unskip}     \fi
\ifx \showLCCN     \undefined \def \showLCCN      #1{\unskip}     \fi
\ifx \shownote     \undefined \def \shownote      #1{#1}          \fi
\ifx \showarticletitle \undefined \def \showarticletitle #1{#1}   \fi
\ifx \showURL      \undefined \def \showURL       {\relax}        \fi
% invisible to TeX
\providecommand\bibfield[2]{#2}
\providecommand\bibinfo[2]{#2}
\providecommand\natexlab[1]{#1}
\providecommand\showeprint[2][]{arXiv:#2}

\bibitem[Int({[n.\,d.]})]%
        {Introduc51:online}
 \bibinfo{year}{[n.\,d.]}\natexlab{}.
\newblock \bibinfo{title}{ChatGPT}.
\newblock \bibinfo{howpublished}{\url{https://openai.com/blog/chatgpt}}.
\newblock
\newblock
\shownote{(Accessed on 11/28/2023)}.


\bibitem[Git({[n.\,d.]})]%
        {GitHubCo19:online}
 \bibinfo{year}{[n.\,d.]}\natexlab{}.
\newblock \bibinfo{title}{GitHub Copilot · Your AI pair programmer}.
\newblock \bibinfo{howpublished}{\url{https://github.com/features/copilot}}.
\newblock
\newblock
\shownote{(Accessed on 11/28/2023)}.


\bibitem[Zer({[n.\,d.]})]%
        {ZeroMQ24:online}
 \bibinfo{year}{[n.\,d.]}\natexlab{}.
\newblock \bibinfo{title}{ZeroMQ}.
\newblock \bibinfo{howpublished}{\url{https://zeromq.org/}}.
\newblock
\newblock
\shownote{(Accessed on 11/29/2023)}.


\bibitem[Alabdulmohsin et~al\mbox{.}(2022)]%
        {alabdulmohsin2022revisiting}
\bibfield{author}{\bibinfo{person}{Ibrahim Alabdulmohsin}, \bibinfo{person}{Behnam Neyshabur}, {and} \bibinfo{person}{Xiaohua Zhai}.} \bibinfo{year}{2022}\natexlab{}.
\newblock \bibinfo{title}{Revisiting Neural Scaling Laws in Language and Vision}.
\newblock
\newblock
\showeprint[arxiv]{2209.06640}~[cs.LG]


\bibitem[Aminabadi et~al\mbox{.}(2022)]%
        {aminabadi2022deepspeed}
\bibfield{author}{\bibinfo{person}{Reza~Yazdani Aminabadi}, \bibinfo{person}{Samyam Rajbhandari}, \bibinfo{person}{Ammar~Ahmad Awan}, \bibinfo{person}{Cheng Li}, \bibinfo{person}{Du Li}, \bibinfo{person}{Elton Zheng}, \bibinfo{person}{Olatunji Ruwase}, \bibinfo{person}{Shaden Smith}, \bibinfo{person}{Minjia Zhang}, \bibinfo{person}{Jeff Rasley}, {et~al\mbox{.}}} \bibinfo{year}{2022}\natexlab{}.
\newblock \showarticletitle{DeepSpeed-inference: enabling efficient inference of transformer models at unprecedented scale}. In \bibinfo{booktitle}{\emph{SC22: International Conference for High Performance Computing, Networking, Storage and Analysis}}. IEEE, \bibinfo{pages}{1--15}.
\newblock


\bibitem[Bai et~al\mbox{.}(2019)]%
        {bai2019}
\bibfield{author}{\bibinfo{person}{Junjie Bai}, \bibinfo{person}{Fang Lu}, \bibinfo{person}{Ke Zhang}, {et~al\mbox{.}}} \bibinfo{year}{2019}\natexlab{}.
\newblock \bibinfo{title}{ONNX: Open Neural Network Exchange}.
\newblock \bibinfo{howpublished}{\url{https://github.com/onnx/onnx}}.
\newblock


\bibitem[Blalock et~al\mbox{.}(2020)]%
        {blalock2020state}
\bibfield{author}{\bibinfo{person}{Davis Blalock}, \bibinfo{person}{Jose~Javier Gonzalez~Ortiz}, \bibinfo{person}{Jonathan Frankle}, {and} \bibinfo{person}{John Guttag}.} \bibinfo{year}{2020}\natexlab{}.
\newblock \showarticletitle{What is the state of neural network pruning?}
\newblock \bibinfo{journal}{\emph{Proceedings of machine learning and systems}}  \bibinfo{volume}{2} (\bibinfo{year}{2020}), \bibinfo{pages}{129--146}.
\newblock


\bibitem[Bondarenko et~al\mbox{.}(2021)]%
        {bondarenko2021understanding}
\bibfield{author}{\bibinfo{person}{Yelysei Bondarenko}, \bibinfo{person}{Markus Nagel}, {and} \bibinfo{person}{Tijmen Blankevoort}.} \bibinfo{year}{2021}\natexlab{}.
\newblock \showarticletitle{Understanding and overcoming the challenges of efficient transformer quantization}.
\newblock \bibinfo{journal}{\emph{arXiv preprint arXiv:2109.12948}} (\bibinfo{year}{2021}).
\newblock


\bibitem[Borzunov et~al\mbox{.}(2022)]%
        {borzunov2022petals}
\bibfield{author}{\bibinfo{person}{Alexander Borzunov}, \bibinfo{person}{Dmitry Baranchuk}, \bibinfo{person}{Tim Dettmers}, \bibinfo{person}{Max Ryabinin}, \bibinfo{person}{Younes Belkada}, \bibinfo{person}{Artem Chumachenko}, \bibinfo{person}{Pavel Samygin}, {and} \bibinfo{person}{Colin Raffel}.} \bibinfo{year}{2022}\natexlab{}.
\newblock \showarticletitle{Petals: Collaborative inference and fine-tuning of large models}.
\newblock \bibinfo{journal}{\emph{arXiv preprint arXiv:2209.01188}} (\bibinfo{year}{2022}).
\newblock


\bibitem[Brown et~al\mbox{.}(2020)]%
        {brown2020language}
\bibfield{author}{\bibinfo{person}{Tom~B. Brown}, \bibinfo{person}{Benjamin Mann}, \bibinfo{person}{Nick Ryder}, \bibinfo{person}{Melanie Subbiah}, \bibinfo{person}{Jared Kaplan}, \bibinfo{person}{Prafulla Dhariwal}, \bibinfo{person}{Arvind Neelakantan}, \bibinfo{person}{Pranav Shyam}, \bibinfo{person}{Girish Sastry}, \bibinfo{person}{Amanda Askell}, \bibinfo{person}{Sandhini Agarwal}, \bibinfo{person}{Ariel Herbert-Voss}, \bibinfo{person}{Gretchen Krueger}, \bibinfo{person}{Tom Henighan}, \bibinfo{person}{Rewon Child}, \bibinfo{person}{Aditya Ramesh}, \bibinfo{person}{Daniel~M. Ziegler}, \bibinfo{person}{Jeffrey Wu}, \bibinfo{person}{Clemens Winter}, \bibinfo{person}{Christopher Hesse}, \bibinfo{person}{Mark Chen}, \bibinfo{person}{Eric Sigler}, \bibinfo{person}{Mateusz Litwin}, \bibinfo{person}{Scott Gray}, \bibinfo{person}{Benjamin Chess}, \bibinfo{person}{Jack Clark}, \bibinfo{person}{Christopher Berner}, \bibinfo{person}{Sam McCandlish}, \bibinfo{person}{Alec Radford}, \bibinfo{person}{Ilya Sutskever},
  {and} \bibinfo{person}{Dario Amodei}.} \bibinfo{year}{2020}\natexlab{}.
\newblock \bibinfo{title}{Language Models are Few-Shot Learners}.
\newblock
\newblock
\showeprint[arxiv]{2005.14165}~[cs.CL]


\bibitem[Chen and Ran(2019)]%
        {chen2019deep}
\bibfield{author}{\bibinfo{person}{Jiasi Chen} {and} \bibinfo{person}{Xukan Ran}.} \bibinfo{year}{2019}\natexlab{}.
\newblock \showarticletitle{Deep learning with edge computing: A review}.
\newblock \bibinfo{journal}{\emph{Proc. IEEE}} \bibinfo{volume}{107}, \bibinfo{number}{8} (\bibinfo{year}{2019}), \bibinfo{pages}{1655--1674}.
\newblock


\bibitem[Chen et~al\mbox{.}(2018)]%
        {chen2018tvm}
\bibfield{author}{\bibinfo{person}{Tianqi Chen}, \bibinfo{person}{Thierry Moreau}, \bibinfo{person}{Ziheng Jiang}, \bibinfo{person}{Lianmin Zheng}, \bibinfo{person}{Eddie Yan}, \bibinfo{person}{Meghan Cowan}, \bibinfo{person}{Haichen Shen}, \bibinfo{person}{Leyuan Wang}, \bibinfo{person}{Yuwei Hu}, \bibinfo{person}{Luis Ceze}, \bibinfo{person}{Carlos Guestrin}, {and} \bibinfo{person}{Arvind Krishnamurthy}.} \bibinfo{year}{2018}\natexlab{}.
\newblock \bibinfo{title}{TVM: An Automated End-to-End Optimizing Compiler for Deep Learning}.
\newblock
\newblock
\showeprint[arxiv]{1802.04799}~[cs.LG]


\bibitem[Chowdhery et~al\mbox{.}(2022)]%
        {chowdhery2022palm}
\bibfield{author}{\bibinfo{person}{Aakanksha Chowdhery}, \bibinfo{person}{Sharan Narang}, \bibinfo{person}{Jacob Devlin}, \bibinfo{person}{Maarten Bosma}, \bibinfo{person}{Gaurav Mishra}, \bibinfo{person}{Adam Roberts}, \bibinfo{person}{Paul Barham}, \bibinfo{person}{Hyung~Won Chung}, \bibinfo{person}{Charles Sutton}, \bibinfo{person}{Sebastian Gehrmann}, \bibinfo{person}{Parker Schuh}, \bibinfo{person}{Kensen Shi}, \bibinfo{person}{Sasha Tsvyashchenko}, \bibinfo{person}{Joshua Maynez}, \bibinfo{person}{Abhishek Rao}, \bibinfo{person}{Parker Barnes}, \bibinfo{person}{Yi Tay}, \bibinfo{person}{Noam Shazeer}, \bibinfo{person}{Vinodkumar Prabhakaran}, \bibinfo{person}{Emily Reif}, \bibinfo{person}{Nan Du}, \bibinfo{person}{Ben Hutchinson}, \bibinfo{person}{Reiner Pope}, \bibinfo{person}{James Bradbury}, \bibinfo{person}{Jacob Austin}, \bibinfo{person}{Michael Isard}, \bibinfo{person}{Guy Gur-Ari}, \bibinfo{person}{Pengcheng Yin}, \bibinfo{person}{Toju Duke}, \bibinfo{person}{Anselm Levskaya},
  \bibinfo{person}{Sanjay Ghemawat}, \bibinfo{person}{Sunipa Dev}, \bibinfo{person}{Henryk Michalewski}, \bibinfo{person}{Xavier Garcia}, \bibinfo{person}{Vedant Misra}, \bibinfo{person}{Kevin Robinson}, \bibinfo{person}{Liam Fedus}, \bibinfo{person}{Denny Zhou}, \bibinfo{person}{Daphne Ippolito}, \bibinfo{person}{David Luan}, \bibinfo{person}{Hyeontaek Lim}, \bibinfo{person}{Barret Zoph}, \bibinfo{person}{Alexander Spiridonov}, \bibinfo{person}{Ryan Sepassi}, \bibinfo{person}{David Dohan}, \bibinfo{person}{Shivani Agrawal}, \bibinfo{person}{Mark Omernick}, \bibinfo{person}{Andrew~M. Dai}, \bibinfo{person}{Thanumalayan~Sankaranarayana Pillai}, \bibinfo{person}{Marie Pellat}, \bibinfo{person}{Aitor Lewkowycz}, \bibinfo{person}{Erica Moreira}, \bibinfo{person}{Rewon Child}, \bibinfo{person}{Oleksandr Polozov}, \bibinfo{person}{Katherine Lee}, \bibinfo{person}{Zongwei Zhou}, \bibinfo{person}{Xuezhi Wang}, \bibinfo{person}{Brennan Saeta}, \bibinfo{person}{Mark Diaz}, \bibinfo{person}{Orhan Firat},
  \bibinfo{person}{Michele Catasta}, \bibinfo{person}{Jason Wei}, \bibinfo{person}{Kathy Meier-Hellstern}, \bibinfo{person}{Douglas Eck}, \bibinfo{person}{Jeff Dean}, \bibinfo{person}{Slav Petrov}, {and} \bibinfo{person}{Noah Fiedel}.} \bibinfo{year}{2022}\natexlab{}.
\newblock \bibinfo{title}{PaLM: Scaling Language Modeling with Pathways}.
\newblock
\newblock
\showeprint[arxiv]{2204.02311}~[cs.CL]


\bibitem[Clark et~al\mbox{.}(2022)]%
        {Clark2022UnifiedSL}
\bibfield{author}{\bibinfo{person}{Aidan Clark}, \bibinfo{person}{Diego de Las~Casas}, \bibinfo{person}{Aurelia Guy}, \bibinfo{person}{Arthur Mensch}, \bibinfo{person}{Michela Paganini}, \bibinfo{person}{Jordan Hoffmann}, \bibinfo{person}{Bogdan Damoc}, \bibinfo{person}{Blake~A. Hechtman}, \bibinfo{person}{Trevor Cai}, \bibinfo{person}{Sebastian Borgeaud}, \bibinfo{person}{George van~den Driessche}, \bibinfo{person}{Eliza Rutherford}, \bibinfo{person}{T.~W. Hennigan}, \bibinfo{person}{Matthew~G. Johnson}, \bibinfo{person}{Katie Millican}, \bibinfo{person}{Albin Cassirer}, \bibinfo{person}{Chris Jones}, \bibinfo{person}{Elena Buchatskaya}, \bibinfo{person}{David Budden}, \bibinfo{person}{L. Sifre}, \bibinfo{person}{Simon Osindero}, \bibinfo{person}{Oriol Vinyals}, \bibinfo{person}{Jack~W. Rae}, \bibinfo{person}{Erich Elsen}, \bibinfo{person}{Koray Kavukcuoglu}, {and} \bibinfo{person}{Karen Simonyan}.} \bibinfo{year}{2022}\natexlab{}.
\newblock \showarticletitle{Unified Scaling Laws for Routed Language Models}. In \bibinfo{booktitle}{\emph{International Conference on Machine Learning}}.
\newblock
\urldef\tempurl%
\url{https://api.semanticscholar.org/CorpusID:246473179}
\showURL{%
\tempurl}


\bibitem[Cobbe et~al\mbox{.}(2021)]%
        {cobbe2021training}
\bibfield{author}{\bibinfo{person}{Karl Cobbe}, \bibinfo{person}{Vineet Kosaraju}, \bibinfo{person}{Mohammad Bavarian}, \bibinfo{person}{Mark Chen}, \bibinfo{person}{Heewoo Jun}, \bibinfo{person}{Lukasz Kaiser}, \bibinfo{person}{Matthias Plappert}, \bibinfo{person}{Jerry Tworek}, \bibinfo{person}{Jacob Hilton}, \bibinfo{person}{Reiichiro Nakano}, \bibinfo{person}{Christopher Hesse}, {and} \bibinfo{person}{John Schulman}.} \bibinfo{year}{2021}\natexlab{}.
\newblock \bibinfo{title}{Training Verifiers to Solve Math Word Problems}.
\newblock
\newblock
\showeprint[arxiv]{2110.14168}~[cs.LG]


\bibitem[Coelho et~al\mbox{.}(2021)]%
        {coelho2021automatic}
\bibfield{author}{\bibinfo{person}{Claudionor~N Coelho}, \bibinfo{person}{Aki Kuusela}, \bibinfo{person}{Shan Li}, \bibinfo{person}{Hao Zhuang}, \bibinfo{person}{Jennifer Ngadiuba}, \bibinfo{person}{Thea~Klaeboe Aarrestad}, \bibinfo{person}{Vladimir Loncar}, \bibinfo{person}{Maurizio Pierini}, \bibinfo{person}{Adrian~Alan Pol}, {and} \bibinfo{person}{Sioni Summers}.} \bibinfo{year}{2021}\natexlab{}.
\newblock \showarticletitle{Automatic heterogeneous quantization of deep neural networks for low-latency inference on the edge for particle detectors}.
\newblock \bibinfo{journal}{\emph{Nature Machine Intelligence}} \bibinfo{volume}{3}, \bibinfo{number}{8} (\bibinfo{year}{2021}), \bibinfo{pages}{675--686}.
\newblock


\bibitem[Du et~al\mbox{.}(2023)]%
        {du2023improving}
\bibfield{author}{\bibinfo{person}{Jiangsu Du}, \bibinfo{person}{Jiazhi Jiang}, \bibinfo{person}{Jiang Zheng}, \bibinfo{person}{Hongbin Zhang}, \bibinfo{person}{Dan Huang}, {and} \bibinfo{person}{Yutong Lu}.} \bibinfo{year}{2023}\natexlab{}.
\newblock \showarticletitle{Improving Computation and Memory Efficiency for Real-world Transformer Inference on GPUs}.
\newblock \bibinfo{journal}{\emph{ACM Transactions on Architecture and Code Optimization}} \bibinfo{volume}{20}, \bibinfo{number}{4} (\bibinfo{year}{2023}), \bibinfo{pages}{1--22}.
\newblock


\bibitem[Elbamby et~al\mbox{.}(2019)]%
        {elbamby2019wireless}
\bibfield{author}{\bibinfo{person}{Mohammed~S Elbamby}, \bibinfo{person}{Cristina Perfecto}, \bibinfo{person}{Chen-Feng Liu}, \bibinfo{person}{Jihong Park}, \bibinfo{person}{Sumudu Samarakoon}, \bibinfo{person}{Xianfu Chen}, {and} \bibinfo{person}{Mehdi Bennis}.} \bibinfo{year}{2019}\natexlab{}.
\newblock \showarticletitle{Wireless edge computing with latency and reliability guarantees}.
\newblock \bibinfo{journal}{\emph{Proc. IEEE}} \bibinfo{volume}{107}, \bibinfo{number}{8} (\bibinfo{year}{2019}), \bibinfo{pages}{1717--1737}.
\newblock


\bibitem[Floridi and Chiriatti(2020)]%
        {gpt3-limit}
\bibfield{author}{\bibinfo{person}{Luciano Floridi} {and} \bibinfo{person}{Massimo Chiriatti}.} \bibinfo{year}{2020}\natexlab{}.
\newblock \showarticletitle{GPT-3: Its Nature, Scope, Limits, and Consequences}.
\newblock \bibinfo{journal}{\emph{Minds and Machines}}  \bibinfo{volume}{30} (\bibinfo{date}{12} \bibinfo{year}{2020}), \bibinfo{pages}{1--14}.
\newblock
\urldef\tempurl%
\url{https://doi.org/10.1007/s11023-020-09548-1}
\showDOI{\tempurl}


\bibitem[Frantar et~al\mbox{.}(2022)]%
        {frantar2022gptq}
\bibfield{author}{\bibinfo{person}{Elias Frantar}, \bibinfo{person}{Saleh Ashkboos}, \bibinfo{person}{Torsten Hoefler}, {and} \bibinfo{person}{Dan Alistarh}.} \bibinfo{year}{2022}\natexlab{}.
\newblock \showarticletitle{Gptq: Accurate post-training quantization for generative pre-trained transformers}.
\newblock \bibinfo{journal}{\emph{arXiv preprint arXiv:2210.17323}} (\bibinfo{year}{2022}).
\newblock


\bibitem[Gholami et~al\mbox{.}(2022)]%
        {gholami2022survey}
\bibfield{author}{\bibinfo{person}{Amir Gholami}, \bibinfo{person}{Sehoon Kim}, \bibinfo{person}{Zhen Dong}, \bibinfo{person}{Zhewei Yao}, \bibinfo{person}{Michael~W Mahoney}, {and} \bibinfo{person}{Kurt Keutzer}.} \bibinfo{year}{2022}\natexlab{}.
\newblock \showarticletitle{A survey of quantization methods for efficient neural network inference}.
\newblock In \bibinfo{booktitle}{\emph{Low-Power Computer Vision}}. \bibinfo{publisher}{Chapman and Hall/CRC}, \bibinfo{pages}{291--326}.
\newblock


\bibitem[Gu et~al\mbox{.}(2023)]%
        {gu2023knowledge}
\bibfield{author}{\bibinfo{person}{Yuxian Gu}, \bibinfo{person}{Li Dong}, \bibinfo{person}{Furu Wei}, {and} \bibinfo{person}{Minlie Huang}.} \bibinfo{year}{2023}\natexlab{}.
\newblock \showarticletitle{Knowledge Distillation of Large Language Models}.
\newblock \bibinfo{journal}{\emph{arXiv preprint arXiv:2306.08543}} (\bibinfo{year}{2023}).
\newblock


\bibitem[Gunasekaran et~al\mbox{.}(2022)]%
        {gunasekaran2022cocktail}
\bibfield{author}{\bibinfo{person}{Jashwant~Raj Gunasekaran}, \bibinfo{person}{Cyan~Subhra Mishra}, \bibinfo{person}{Prashanth Thinakaran}, \bibinfo{person}{Bikash Sharma}, \bibinfo{person}{Mahmut~Taylan Kandemir}, {and} \bibinfo{person}{Chita~R Das}.} \bibinfo{year}{2022}\natexlab{}.
\newblock \showarticletitle{Cocktail: A multidimensional optimization for model serving in cloud}. In \bibinfo{booktitle}{\emph{19th USENIX Symposium on Networked Systems Design and Implementation (NSDI 22)}}. \bibinfo{pages}{1041--1057}.
\newblock


\bibitem[Guo(2018)]%
        {guo2018survey}
\bibfield{author}{\bibinfo{person}{Yunhui Guo}.} \bibinfo{year}{2018}\natexlab{}.
\newblock \showarticletitle{A survey on methods and theories of quantized neural networks}.
\newblock \bibinfo{journal}{\emph{arXiv preprint arXiv:1808.04752}} (\bibinfo{year}{2018}).
\newblock


\bibitem[{Gurobi Optimization, LLC}(2023)]%
        {gurobi}
\bibfield{author}{\bibinfo{person}{{Gurobi Optimization, LLC}}.} \bibinfo{year}{2023}\natexlab{}.
\newblock \bibinfo{title}{{Gurobi Optimizer Reference Manual}}.
\newblock
\newblock
\urldef\tempurl%
\url{https://www.gurobi.com}
\showURL{%
\tempurl}


\bibitem[Hendrycks et~al\mbox{.}(2021)]%
        {hendrycks2021measuring}
\bibfield{author}{\bibinfo{person}{Dan Hendrycks}, \bibinfo{person}{Collin Burns}, \bibinfo{person}{Steven Basart}, \bibinfo{person}{Andy Zou}, \bibinfo{person}{Mantas Mazeika}, \bibinfo{person}{Dawn Song}, {and} \bibinfo{person}{Jacob Steinhardt}.} \bibinfo{year}{2021}\natexlab{}.
\newblock \bibinfo{title}{Measuring Massive Multitask Language Understanding}.
\newblock
\newblock
\showeprint[arxiv]{2009.03300}~[cs.CY]


\bibitem[Hoefler et~al\mbox{.}(2021)]%
        {hoefler2021sparsity}
\bibfield{author}{\bibinfo{person}{Torsten Hoefler}, \bibinfo{person}{Dan Alistarh}, \bibinfo{person}{Tal Ben-Nun}, \bibinfo{person}{Nikoli Dryden}, {and} \bibinfo{person}{Alexandra Peste}.} \bibinfo{year}{2021}\natexlab{}.
\newblock \showarticletitle{Sparsity in deep learning: Pruning and growth for efficient inference and training in neural networks}.
\newblock \bibinfo{journal}{\emph{The Journal of Machine Learning Research}} \bibinfo{volume}{22}, \bibinfo{number}{1} (\bibinfo{year}{2021}), \bibinfo{pages}{10882--11005}.
\newblock


\bibitem[Hoffmann et~al\mbox{.}(2022)]%
        {hoffmann2022training}
\bibfield{author}{\bibinfo{person}{Jordan Hoffmann}, \bibinfo{person}{Sebastian Borgeaud}, \bibinfo{person}{Arthur Mensch}, \bibinfo{person}{Elena Buchatskaya}, \bibinfo{person}{Trevor Cai}, \bibinfo{person}{Eliza Rutherford}, \bibinfo{person}{Diego de Las~Casas}, \bibinfo{person}{Lisa~Anne Hendricks}, \bibinfo{person}{Johannes Welbl}, \bibinfo{person}{Aidan Clark}, \bibinfo{person}{Tom Hennigan}, \bibinfo{person}{Eric Noland}, \bibinfo{person}{Katie Millican}, \bibinfo{person}{George van~den Driessche}, \bibinfo{person}{Bogdan Damoc}, \bibinfo{person}{Aurelia Guy}, \bibinfo{person}{Simon Osindero}, \bibinfo{person}{Karen Simonyan}, \bibinfo{person}{Erich Elsen}, \bibinfo{person}{Jack~W. Rae}, \bibinfo{person}{Oriol Vinyals}, {and} \bibinfo{person}{Laurent Sifre}.} \bibinfo{year}{2022}\natexlab{}.
\newblock \bibinfo{title}{Training Compute-Optimal Large Language Models}.
\newblock
\newblock
\showeprint[arxiv]{2203.15556}~[cs.CL]


\bibitem[Hu and Li(2022)]%
        {hu2022distributed}
\bibfield{author}{\bibinfo{person}{Chenghao Hu} {and} \bibinfo{person}{Baochun Li}.} \bibinfo{year}{2022}\natexlab{}.
\newblock \showarticletitle{Distributed inference with deep learning models across heterogeneous edge devices}. In \bibinfo{booktitle}{\emph{IEEE INFOCOM 2022-IEEE Conference on Computer Communications}}. IEEE, \bibinfo{pages}{330--339}.
\newblock


\bibitem[Hu et~al\mbox{.}(2019)]%
        {hu2019deephome}
\bibfield{author}{\bibinfo{person}{Zhiming Hu}, \bibinfo{person}{Ahmad~Bisher Tarakji}, \bibinfo{person}{Vishal Raheja}, \bibinfo{person}{Caleb Phillips}, \bibinfo{person}{Teng Wang}, {and} \bibinfo{person}{Iqbal Mohomed}.} \bibinfo{year}{2019}\natexlab{}.
\newblock \showarticletitle{Deephome: Distributed inference with heterogeneous devices in the edge}. In \bibinfo{booktitle}{\emph{The 3rd International Workshop on Deep Learning for Mobile Systems and Applications}}. \bibinfo{pages}{13--18}.
\newblock


\bibitem[Huang et~al\mbox{.}(2020)]%
        {10.1145/3373376.3378530}
\bibfield{author}{\bibinfo{person}{Chien-Chin Huang}, \bibinfo{person}{Gu Jin}, {and} \bibinfo{person}{Jinyang Li}.} \bibinfo{year}{2020}\natexlab{}.
\newblock \showarticletitle{SwapAdvisor: Pushing Deep Learning Beyond the GPU Memory Limit via Smart Swapping}. In \bibinfo{booktitle}{\emph{Proceedings of the Twenty-Fifth International Conference on Architectural Support for Programming Languages and Operating Systems}} (Lausanne, Switzerland) \emph{(\bibinfo{series}{ASPLOS '20})}. \bibinfo{publisher}{Association for Computing Machinery}, \bibinfo{address}{New York, NY, USA}, \bibinfo{pages}{1341–1355}.
\newblock
\showISBNx{9781450371025}
\urldef\tempurl%
\url{https://doi.org/10.1145/3373376.3378530}
\showDOI{\tempurl}


\bibitem[Jiao et~al\mbox{.}(2019)]%
        {jiao2019tinybert}
\bibfield{author}{\bibinfo{person}{Xiaoqi Jiao}, \bibinfo{person}{Yichun Yin}, \bibinfo{person}{Lifeng Shang}, \bibinfo{person}{Xin Jiang}, \bibinfo{person}{Xiao Chen}, \bibinfo{person}{Linlin Li}, \bibinfo{person}{Fang Wang}, {and} \bibinfo{person}{Qun Liu}.} \bibinfo{year}{2019}\natexlab{}.
\newblock \showarticletitle{Tinybert: Distilling bert for natural language understanding}.
\newblock \bibinfo{journal}{\emph{arXiv preprint arXiv:1909.10351}} (\bibinfo{year}{2019}).
\newblock


\bibitem[Kaplan et~al\mbox{.}(2020)]%
        {kaplan2020scaling}
\bibfield{author}{\bibinfo{person}{Jared Kaplan}, \bibinfo{person}{Sam McCandlish}, \bibinfo{person}{Tom Henighan}, \bibinfo{person}{Tom~B. Brown}, \bibinfo{person}{Benjamin Chess}, \bibinfo{person}{Rewon Child}, \bibinfo{person}{Scott Gray}, \bibinfo{person}{Alec Radford}, \bibinfo{person}{Jeffrey Wu}, {and} \bibinfo{person}{Dario Amodei}.} \bibinfo{year}{2020}\natexlab{}.
\newblock \bibinfo{title}{Scaling Laws for Neural Language Models}.
\newblock
\newblock
\showeprint[arxiv]{2001.08361}~[cs.LG]


\bibitem[Khowaja et~al\mbox{.}(2023)]%
        {khowaja2023chatgpt}
\bibfield{author}{\bibinfo{person}{Sunder~Ali Khowaja}, \bibinfo{person}{Parus Khuwaja}, {and} \bibinfo{person}{Kapal Dev}.} \bibinfo{year}{2023}\natexlab{}.
\newblock \showarticletitle{ChatGPT Needs SPADE (Sustainability, PrivAcy, Digital divide, and Ethics) Evaluation: A Review}.
\newblock \bibinfo{journal}{\emph{arXiv preprint arXiv:2305.03123}} (\bibinfo{year}{2023}).
\newblock


\bibitem[Kshetri(2023)]%
        {kshetri2023cybercrime}
\bibfield{author}{\bibinfo{person}{Nir Kshetri}.} \bibinfo{year}{2023}\natexlab{}.
\newblock \showarticletitle{Cybercrime and Privacy Threats of Large Language Models}.
\newblock \bibinfo{journal}{\emph{IT Professional}} \bibinfo{volume}{25}, \bibinfo{number}{3} (\bibinfo{year}{2023}), \bibinfo{pages}{9--13}.
\newblock


\bibitem[Lee(2023)]%
        {math11112451}
\bibfield{author}{\bibinfo{person}{Minhyeok Lee}.} \bibinfo{year}{2023}\natexlab{}.
\newblock \showarticletitle{A Mathematical Interpretation of Autoregressive Generative Pre-Trained Transformer and Self-Supervised Learning}.
\newblock \bibinfo{journal}{\emph{Mathematics}} \bibinfo{volume}{11}, \bibinfo{number}{11} (\bibinfo{year}{2023}).
\newblock
\showISSN{2227-7390}
\urldef\tempurl%
\url{https://doi.org/10.3390/math11112451}
\showDOI{\tempurl}


\bibitem[Liang et~al\mbox{.}(2020b)]%
        {liang2020toward}
\bibfield{author}{\bibinfo{person}{Fan Liang}, \bibinfo{person}{Wei Yu}, \bibinfo{person}{Xing Liu}, \bibinfo{person}{David Griffith}, {and} \bibinfo{person}{Nada Golmie}.} \bibinfo{year}{2020}\natexlab{b}.
\newblock \showarticletitle{Toward edge-based deep learning in industrial Internet of Things}.
\newblock \bibinfo{journal}{\emph{IEEE Internet of Things Journal}} \bibinfo{volume}{7}, \bibinfo{number}{5} (\bibinfo{year}{2020}), \bibinfo{pages}{4329--4341}.
\newblock


\bibitem[Liang et~al\mbox{.}(2020a)]%
        {liang2020mixkd}
\bibfield{author}{\bibinfo{person}{Kevin~J Liang}, \bibinfo{person}{Weituo Hao}, \bibinfo{person}{Dinghan Shen}, \bibinfo{person}{Yufan Zhou}, \bibinfo{person}{Weizhu Chen}, \bibinfo{person}{Changyou Chen}, {and} \bibinfo{person}{Lawrence Carin}.} \bibinfo{year}{2020}\natexlab{a}.
\newblock \showarticletitle{Mixkd: Towards efficient distillation of large-scale language models}.
\newblock \bibinfo{journal}{\emph{arXiv preprint arXiv:2011.00593}} (\bibinfo{year}{2020}).
\newblock


\bibitem[Liang et~al\mbox{.}(2021)]%
        {liang2021pruning}
\bibfield{author}{\bibinfo{person}{Tailin Liang}, \bibinfo{person}{John Glossner}, \bibinfo{person}{Lei Wang}, \bibinfo{person}{Shaobo Shi}, {and} \bibinfo{person}{Xiaotong Zhang}.} \bibinfo{year}{2021}\natexlab{}.
\newblock \showarticletitle{Pruning and quantization for deep neural network acceleration: A survey}.
\newblock \bibinfo{journal}{\emph{Neurocomputing}}  \bibinfo{volume}{461} (\bibinfo{year}{2021}), \bibinfo{pages}{370--403}.
\newblock


\bibitem[Lin et~al\mbox{.}(2021)]%
        {lin2021limitations}
\bibfield{author}{\bibinfo{person}{Chu-Cheng Lin}, \bibinfo{person}{Aaron Jaech}, \bibinfo{person}{Xin Li}, \bibinfo{person}{Matthew~R. Gormley}, {and} \bibinfo{person}{Jason Eisner}.} \bibinfo{year}{2021}\natexlab{}.
\newblock \bibinfo{title}{Limitations of Autoregressive Models and Their Alternatives}.
\newblock
\newblock
\showeprint[arxiv]{2010.11939}~[cs.LG]


\bibitem[Maas et~al\mbox{.}(2011)]%
        {maas-EtAl:2011:ACL-HLT2011}
\bibfield{author}{\bibinfo{person}{Andrew~L. Maas}, \bibinfo{person}{Raymond~E. Daly}, \bibinfo{person}{Peter~T. Pham}, \bibinfo{person}{Dan Huang}, \bibinfo{person}{Andrew~Y. Ng}, {and} \bibinfo{person}{Christopher Potts}.} \bibinfo{year}{2011}\natexlab{}.
\newblock \showarticletitle{Learning Word Vectors for Sentiment Analysis}. In \bibinfo{booktitle}{\emph{Proceedings of the 49th Annual Meeting of the Association for Computational Linguistics: Human Language Technologies}}. \bibinfo{publisher}{Association for Computational Linguistics}, \bibinfo{address}{Portland, Oregon, USA}, \bibinfo{pages}{142--150}.
\newblock
\urldef\tempurl%
\url{http://www.aclweb.org/anthology/P11-1015}
\showURL{%
\tempurl}


\bibitem[Mao et~al\mbox{.}(2017a)]%
        {mao2017modnn}
\bibfield{author}{\bibinfo{person}{Jiachen Mao}, \bibinfo{person}{Xiang Chen}, \bibinfo{person}{Kent~W Nixon}, \bibinfo{person}{Christopher Krieger}, {and} \bibinfo{person}{Yiran Chen}.} \bibinfo{year}{2017}\natexlab{a}.
\newblock \showarticletitle{Modnn: Local distributed mobile computing system for deep neural network}. In \bibinfo{booktitle}{\emph{Design, Automation \& Test in Europe Conference \& Exhibition (DATE), 2017}}. IEEE, \bibinfo{pages}{1396--1401}.
\newblock


\bibitem[Mao et~al\mbox{.}(2017b)]%
        {mao2017survey}
\bibfield{author}{\bibinfo{person}{Yuyi Mao}, \bibinfo{person}{Changsheng You}, \bibinfo{person}{Jun Zhang}, \bibinfo{person}{Kaibin Huang}, {and} \bibinfo{person}{Khaled~B Letaief}.} \bibinfo{year}{2017}\natexlab{b}.
\newblock \showarticletitle{A survey on mobile edge computing: The communication perspective}.
\newblock \bibinfo{journal}{\emph{IEEE communications surveys \& tutorials}} \bibinfo{volume}{19}, \bibinfo{number}{4} (\bibinfo{year}{2017}), \bibinfo{pages}{2322--2358}.
\newblock


\bibitem[Merity et~al\mbox{.}(2016)]%
        {merity2016pointer}
\bibfield{author}{\bibinfo{person}{Stephen Merity}, \bibinfo{person}{Caiming Xiong}, \bibinfo{person}{James Bradbury}, {and} \bibinfo{person}{Richard Socher}.} \bibinfo{year}{2016}\natexlab{}.
\newblock \bibinfo{title}{Pointer Sentinel Mixture Models}.
\newblock
\newblock
\showeprint[arxiv]{1609.07843}~[cs.CL]


\bibitem[Naveen et~al\mbox{.}(2021)]%
        {naveen2021low}
\bibfield{author}{\bibinfo{person}{Soumyalatha Naveen}, \bibinfo{person}{Manjunath~R Kounte}, {and} \bibinfo{person}{Mohammed~Riyaz Ahmed}.} \bibinfo{year}{2021}\natexlab{}.
\newblock \showarticletitle{Low latency deep learning inference model for distributed intelligent IoT edge clusters}.
\newblock \bibinfo{journal}{\emph{IEEE Access}}  \bibinfo{volume}{9} (\bibinfo{year}{2021}), \bibinfo{pages}{160607--160621}.
\newblock


\bibitem[Park et~al\mbox{.}(2019)]%
        {park2019wireless}
\bibfield{author}{\bibinfo{person}{Jihong Park}, \bibinfo{person}{Sumudu Samarakoon}, \bibinfo{person}{Mehdi Bennis}, {and} \bibinfo{person}{M{\'e}rouane Debbah}.} \bibinfo{year}{2019}\natexlab{}.
\newblock \showarticletitle{Wireless network intelligence at the edge}.
\newblock \bibinfo{journal}{\emph{Proc. IEEE}} \bibinfo{volume}{107}, \bibinfo{number}{11} (\bibinfo{year}{2019}), \bibinfo{pages}{2204--2239}.
\newblock


\bibitem[Paszke et~al\mbox{.}(2019)]%
        {paszke2019pytorch}
\bibfield{author}{\bibinfo{person}{Adam Paszke}, \bibinfo{person}{Sam Gross}, \bibinfo{person}{Francisco Massa}, \bibinfo{person}{Adam Lerer}, \bibinfo{person}{James Bradbury}, \bibinfo{person}{Gregory Chanan}, \bibinfo{person}{Trevor Killeen}, \bibinfo{person}{Zeming Lin}, \bibinfo{person}{Natalia Gimelshein}, \bibinfo{person}{Luca Antiga}, \bibinfo{person}{Alban Desmaison}, \bibinfo{person}{Andreas Köpf}, \bibinfo{person}{Edward Yang}, \bibinfo{person}{Zach DeVito}, \bibinfo{person}{Martin Raison}, \bibinfo{person}{Alykhan Tejani}, \bibinfo{person}{Sasank Chilamkurthy}, \bibinfo{person}{Benoit Steiner}, \bibinfo{person}{Lu Fang}, \bibinfo{person}{Junjie Bai}, {and} \bibinfo{person}{Soumith Chintala}.} \bibinfo{year}{2019}\natexlab{}.
\newblock \bibinfo{title}{PyTorch: An Imperative Style, High-Performance Deep Learning Library}.
\newblock
\newblock
\showeprint[arxiv]{1912.01703}~[cs.LG]


\bibitem[Patel and Pavlick(2022)]%
        {patel2022mapping}
\bibfield{author}{\bibinfo{person}{Roma Patel} {and} \bibinfo{person}{Ellie Pavlick}.} \bibinfo{year}{2022}\natexlab{}.
\newblock \showarticletitle{Mapping Language Models to Grounded Conceptual Spaces}. In \bibinfo{booktitle}{\emph{International Conference on Learning Representations}}.
\newblock
\urldef\tempurl%
\url{https://openreview.net/forum?id=gJcEM8sxHK}
\showURL{%
\tempurl}


\bibitem[Peng et~al\mbox{.}(2023)]%
        {peng2023instruction}
\bibfield{author}{\bibinfo{person}{Baolin Peng}, \bibinfo{person}{Chunyuan Li}, \bibinfo{person}{Pengcheng He}, \bibinfo{person}{Michel Galley}, {and} \bibinfo{person}{Jianfeng Gao}.} \bibinfo{year}{2023}\natexlab{}.
\newblock \bibinfo{title}{Instruction Tuning with GPT-4}.
\newblock
\newblock
\showeprint[arxiv]{2304.03277}~[cs.CL]


\bibitem[Rasley et~al\mbox{.}(2020)]%
        {rasley2020deepspeed}
\bibfield{author}{\bibinfo{person}{Jeff Rasley}, \bibinfo{person}{Samyam Rajbhandari}, \bibinfo{person}{Olatunji Ruwase}, {and} \bibinfo{person}{Yuxiong He}.} \bibinfo{year}{2020}\natexlab{}.
\newblock \showarticletitle{Deepspeed: System optimizations enable training deep learning models with over 100 billion parameters}. In \bibinfo{booktitle}{\emph{Proceedings of the 26th ACM SIGKDD International Conference on Knowledge Discovery \& Data Mining}}. \bibinfo{pages}{3505--3506}.
\newblock


\bibitem[Reed et~al\mbox{.}(2022)]%
        {reed2022torchfx}
\bibfield{author}{\bibinfo{person}{James~K. Reed}, \bibinfo{person}{Zachary DeVito}, \bibinfo{person}{Horace He}, \bibinfo{person}{Ansley Ussery}, {and} \bibinfo{person}{Jason Ansel}.} \bibinfo{year}{2022}\natexlab{}.
\newblock \bibinfo{title}{Torch.fx: Practical Program Capture and Transformation for Deep Learning in Python}.
\newblock
\newblock
\showeprint[arxiv]{2112.08429}~[cs.LG]


\bibitem[Renaud et~al\mbox{.}(2023)]%
        {renaud2023chatgpt}
\bibfield{author}{\bibinfo{person}{Karen Renaud}, \bibinfo{person}{Merrill Warkentin}, {and} \bibinfo{person}{George Westerman}.} \bibinfo{year}{2023}\natexlab{}.
\newblock \bibinfo{booktitle}{\emph{From ChatGPT to HackGPT: Meeting the Cybersecurity Threat of Generative AI}}.
\newblock \bibinfo{publisher}{MIT Sloan Management Review}.
\newblock


\bibitem[Romero et~al\mbox{.}(2021)]%
        {romero2021infaas}
\bibfield{author}{\bibinfo{person}{Francisco Romero}, \bibinfo{person}{Qian Li}, \bibinfo{person}{Neeraja~J Yadwadkar}, {and} \bibinfo{person}{Christos Kozyrakis}.} \bibinfo{year}{2021}\natexlab{}.
\newblock \showarticletitle{$\{$INFaaS$\}$: Automated model-less inference serving}. In \bibinfo{booktitle}{\emph{2021 USENIX Annual Technical Conference (USENIX ATC 21)}}. \bibinfo{pages}{397--411}.
\newblock


\bibitem[Runtime(2023)]%
        {ONNXRuntime}
\bibfield{author}{\bibinfo{person}{ONNX Runtime}.} \bibinfo{year}{2023}\natexlab{}.
\newblock \bibinfo{title}{ONNX Runtime}.
\newblock
\newblock
\newblock
\shownote{Available from: \url{https://onnxruntime.ai/}}.


\bibitem[Sebastian(2023)]%
        {sebastian2023chatgpt}
\bibfield{author}{\bibinfo{person}{Glorin Sebastian}.} \bibinfo{year}{2023}\natexlab{}.
\newblock \showarticletitle{Do ChatGPT and other AI chatbots pose a cybersecurity risk?: An exploratory study}.
\newblock \bibinfo{journal}{\emph{International Journal of Security and Privacy in Pervasive Computing (IJSPPC)}} \bibinfo{volume}{15}, \bibinfo{number}{1} (\bibinfo{year}{2023}), \bibinfo{pages}{1--11}.
\newblock


\bibitem[team(2023)]%
        {mlc-llm}
\bibfield{author}{\bibinfo{person}{MLC team}.} \bibinfo{year}{2023}\natexlab{}.
\newblock \bibinfo{booktitle}{\emph{{MLC-LLM}}}.
\newblock
\urldef\tempurl%
\url{https://github.com/mlc-ai/mlc-llm}
\showURL{%
\tempurl}


\bibitem[TensorFlow(2023)]%
        {TensorFlowLite}
\bibfield{author}{\bibinfo{person}{TensorFlow}.} \bibinfo{year}{2023}\natexlab{}.
\newblock \bibinfo{title}{TensorFlow Lite}.
\newblock
\newblock
\newblock
\shownote{Available from: \url{https://www.tensorflow.org/lite}}.


\bibitem[Touvron et~al\mbox{.}(2023)]%
        {touvron2023llama}
\bibfield{author}{\bibinfo{person}{Hugo Touvron}, \bibinfo{person}{Thibaut Lavril}, \bibinfo{person}{Gautier Izacard}, \bibinfo{person}{Xavier Martinet}, \bibinfo{person}{Marie-Anne Lachaux}, \bibinfo{person}{Timothée Lacroix}, \bibinfo{person}{Baptiste Rozière}, \bibinfo{person}{Naman Goyal}, \bibinfo{person}{Eric Hambro}, \bibinfo{person}{Faisal Azhar}, \bibinfo{person}{Aurelien Rodriguez}, \bibinfo{person}{Armand Joulin}, \bibinfo{person}{Edouard Grave}, {and} \bibinfo{person}{Guillaume Lample}.} \bibinfo{year}{2023}\natexlab{}.
\newblock \bibinfo{title}{LLaMA: Open and Efficient Foundation Language Models}.
\newblock
\newblock
\showeprint[arxiv]{2302.13971}~[cs.CL]


\bibitem[Vaswani et~al\mbox{.}(2023)]%
        {vaswani2023attention}
\bibfield{author}{\bibinfo{person}{Ashish Vaswani}, \bibinfo{person}{Noam Shazeer}, \bibinfo{person}{Niki Parmar}, \bibinfo{person}{Jakob Uszkoreit}, \bibinfo{person}{Llion Jones}, \bibinfo{person}{Aidan~N. Gomez}, \bibinfo{person}{Lukasz Kaiser}, {and} \bibinfo{person}{Illia Polosukhin}.} \bibinfo{year}{2023}\natexlab{}.
\newblock \bibinfo{title}{Attention Is All You Need}.
\newblock
\newblock
\showeprint[arxiv]{1706.03762}~[cs.CL]


\bibitem[Wang et~al\mbox{.}(2023)]%
        {wang2023tabi}
\bibfield{author}{\bibinfo{person}{Yiding Wang}, \bibinfo{person}{Kai Chen}, \bibinfo{person}{Haisheng Tan}, {and} \bibinfo{person}{Kun Guo}.} \bibinfo{year}{2023}\natexlab{}.
\newblock \showarticletitle{Tabi: An Efficient Multi-Level Inference System for Large Language Models}. In \bibinfo{booktitle}{\emph{Proceedings of the Eighteenth European Conference on Computer Systems}}. \bibinfo{pages}{233--248}.
\newblock


\bibitem[Workshop et~al\mbox{.}(2023)]%
        {workshop2023bloom}
\bibfield{author}{\bibinfo{person}{BigScience Workshop}, \bibinfo{person}{:}, \bibinfo{person}{Teven~Le Scao}, \bibinfo{person}{Angela Fan}, \bibinfo{person}{Christopher Akiki}, \bibinfo{person}{Ellie Pavlick}, \bibinfo{person}{Suzana Ilić}, \bibinfo{person}{Daniel Hesslow}, \bibinfo{person}{Roman Castagné}, \bibinfo{person}{Alexandra~Sasha Luccioni}, \bibinfo{person}{François Yvon}, \bibinfo{person}{Matthias Gallé}, \bibinfo{person}{Jonathan Tow}, \bibinfo{person}{Alexander~M. Rush}, \bibinfo{person}{Stella Biderman}, \bibinfo{person}{Albert Webson}, \bibinfo{person}{Pawan~Sasanka Ammanamanchi}, \bibinfo{person}{Thomas Wang}, \bibinfo{person}{Benoît Sagot}, \bibinfo{person}{Niklas Muennighoff}, \bibinfo{person}{Albert~Villanova del Moral}, \bibinfo{person}{Olatunji Ruwase}, \bibinfo{person}{Rachel Bawden}, \bibinfo{person}{Stas Bekman}, \bibinfo{person}{Angelina McMillan-Major}, \bibinfo{person}{Iz Beltagy}, \bibinfo{person}{Huu Nguyen}, \bibinfo{person}{Lucile Saulnier}, \bibinfo{person}{Samson Tan},
  \bibinfo{person}{Pedro~Ortiz Suarez}, \bibinfo{person}{Victor Sanh}, \bibinfo{person}{Hugo Laurençon}, \bibinfo{person}{Yacine Jernite}, \bibinfo{person}{Julien Launay}, \bibinfo{person}{Margaret Mitchell}, \bibinfo{person}{Colin Raffel}, \bibinfo{person}{Aaron Gokaslan}, \bibinfo{person}{Adi Simhi}, \bibinfo{person}{Aitor Soroa}, \bibinfo{person}{Alham~Fikri Aji}, \bibinfo{person}{Amit Alfassy}, \bibinfo{person}{Anna Rogers}, \bibinfo{person}{Ariel~Kreisberg Nitzav}, \bibinfo{person}{Canwen Xu}, \bibinfo{person}{Chenghao Mou}, \bibinfo{person}{Chris Emezue}, \bibinfo{person}{Christopher Klamm}, \bibinfo{person}{Colin Leong}, \bibinfo{person}{Daniel van Strien}, \bibinfo{person}{David~Ifeoluwa Adelani}, \bibinfo{person}{Dragomir Radev}, \bibinfo{person}{Eduardo~González Ponferrada}, \bibinfo{person}{Efrat Levkovizh}, \bibinfo{person}{Ethan Kim}, \bibinfo{person}{Eyal~Bar Natan}, \bibinfo{person}{Francesco~De Toni}, \bibinfo{person}{Gérard Dupont}, \bibinfo{person}{Germán Kruszewski},
  \bibinfo{person}{Giada Pistilli}, \bibinfo{person}{Hady Elsahar}, \bibinfo{person}{Hamza Benyamina}, \bibinfo{person}{Hieu Tran}, \bibinfo{person}{Ian Yu}, \bibinfo{person}{Idris Abdulmumin}, \bibinfo{person}{Isaac Johnson}, \bibinfo{person}{Itziar Gonzalez-Dios}, \bibinfo{person}{Javier de~la Rosa}, \bibinfo{person}{Jenny Chim}, \bibinfo{person}{Jesse Dodge}, \bibinfo{person}{Jian Zhu}, \bibinfo{person}{Jonathan Chang}, \bibinfo{person}{Jörg Frohberg}, \bibinfo{person}{Joseph Tobing}, \bibinfo{person}{Joydeep Bhattacharjee}, \bibinfo{person}{Khalid Almubarak}, \bibinfo{person}{Kimbo Chen}, \bibinfo{person}{Kyle Lo}, \bibinfo{person}{Leandro~Von Werra}, \bibinfo{person}{Leon Weber}, \bibinfo{person}{Long Phan}, \bibinfo{person}{Loubna~Ben allal}, \bibinfo{person}{Ludovic Tanguy}, \bibinfo{person}{Manan Dey}, \bibinfo{person}{Manuel~Romero Muñoz}, \bibinfo{person}{Maraim Masoud}, \bibinfo{person}{María Grandury}, \bibinfo{person}{Mario Šaško}, \bibinfo{person}{Max Huang}, \bibinfo{person}{Maximin
  Coavoux}, \bibinfo{person}{Mayank Singh}, \bibinfo{person}{Mike Tian-Jian Jiang}, \bibinfo{person}{Minh~Chien Vu}, \bibinfo{person}{Mohammad~A. Jauhar}, \bibinfo{person}{Mustafa Ghaleb}, \bibinfo{person}{Nishant Subramani}, \bibinfo{person}{Nora Kassner}, \bibinfo{person}{Nurulaqilla Khamis}, \bibinfo{person}{Olivier Nguyen}, \bibinfo{person}{Omar Espejel}, \bibinfo{person}{Ona de Gibert}, \bibinfo{person}{Paulo Villegas}, \bibinfo{person}{Peter Henderson}, \bibinfo{person}{Pierre Colombo}, \bibinfo{person}{Priscilla Amuok}, \bibinfo{person}{Quentin Lhoest}, \bibinfo{person}{Rheza Harliman}, \bibinfo{person}{Rishi Bommasani}, \bibinfo{person}{Roberto~Luis López}, \bibinfo{person}{Rui Ribeiro}, \bibinfo{person}{Salomey Osei}, \bibinfo{person}{Sampo Pyysalo}, \bibinfo{person}{Sebastian Nagel}, \bibinfo{person}{Shamik Bose}, \bibinfo{person}{Shamsuddeen~Hassan Muhammad}, \bibinfo{person}{Shanya Sharma}, \bibinfo{person}{Shayne Longpre}, \bibinfo{person}{Somaieh Nikpoor}, \bibinfo{person}{Stanislav
  Silberberg}, \bibinfo{person}{Suhas Pai}, \bibinfo{person}{Sydney Zink}, \bibinfo{person}{Tiago~Timponi Torrent}, \bibinfo{person}{Timo Schick}, \bibinfo{person}{Tristan Thrush}, \bibinfo{person}{Valentin Danchev}, \bibinfo{person}{Vassilina Nikoulina}, \bibinfo{person}{Veronika Laippala}, \bibinfo{person}{Violette Lepercq}, \bibinfo{person}{Vrinda Prabhu}, \bibinfo{person}{Zaid Alyafeai}, \bibinfo{person}{Zeerak Talat}, \bibinfo{person}{Arun Raja}, \bibinfo{person}{Benjamin Heinzerling}, \bibinfo{person}{Chenglei Si}, \bibinfo{person}{Davut~Emre Taşar}, \bibinfo{person}{Elizabeth Salesky}, \bibinfo{person}{Sabrina~J. Mielke}, \bibinfo{person}{Wilson~Y. Lee}, \bibinfo{person}{Abheesht Sharma}, \bibinfo{person}{Andrea Santilli}, \bibinfo{person}{Antoine Chaffin}, \bibinfo{person}{Arnaud Stiegler}, \bibinfo{person}{Debajyoti Datta}, \bibinfo{person}{Eliza Szczechla}, \bibinfo{person}{Gunjan Chhablani}, \bibinfo{person}{Han Wang}, \bibinfo{person}{Harshit Pandey}, \bibinfo{person}{Hendrik Strobelt},
  \bibinfo{person}{Jason~Alan Fries}, \bibinfo{person}{Jos Rozen}, \bibinfo{person}{Leo Gao}, \bibinfo{person}{Lintang Sutawika}, \bibinfo{person}{M~Saiful Bari}, \bibinfo{person}{Maged~S. Al-shaibani}, \bibinfo{person}{Matteo Manica}, \bibinfo{person}{Nihal Nayak}, \bibinfo{person}{Ryan Teehan}, \bibinfo{person}{Samuel Albanie}, \bibinfo{person}{Sheng Shen}, \bibinfo{person}{Srulik Ben-David}, \bibinfo{person}{Stephen~H. Bach}, \bibinfo{person}{Taewoon Kim}, \bibinfo{person}{Tali Bers}, \bibinfo{person}{Thibault Fevry}, \bibinfo{person}{Trishala Neeraj}, \bibinfo{person}{Urmish Thakker}, \bibinfo{person}{Vikas Raunak}, \bibinfo{person}{Xiangru Tang}, \bibinfo{person}{Zheng-Xin Yong}, \bibinfo{person}{Zhiqing Sun}, \bibinfo{person}{Shaked Brody}, \bibinfo{person}{Yallow Uri}, \bibinfo{person}{Hadar Tojarieh}, \bibinfo{person}{Adam Roberts}, \bibinfo{person}{Hyung~Won Chung}, \bibinfo{person}{Jaesung Tae}, \bibinfo{person}{Jason Phang}, \bibinfo{person}{Ofir Press}, \bibinfo{person}{Conglong Li},
  \bibinfo{person}{Deepak Narayanan}, \bibinfo{person}{Hatim Bourfoune}, \bibinfo{person}{Jared Casper}, \bibinfo{person}{Jeff Rasley}, \bibinfo{person}{Max Ryabinin}, \bibinfo{person}{Mayank Mishra}, \bibinfo{person}{Minjia Zhang}, \bibinfo{person}{Mohammad Shoeybi}, \bibinfo{person}{Myriam Peyrounette}, \bibinfo{person}{Nicolas Patry}, \bibinfo{person}{Nouamane Tazi}, \bibinfo{person}{Omar Sanseviero}, \bibinfo{person}{Patrick von Platen}, \bibinfo{person}{Pierre Cornette}, \bibinfo{person}{Pierre~François Lavallée}, \bibinfo{person}{Rémi Lacroix}, \bibinfo{person}{Samyam Rajbhandari}, \bibinfo{person}{Sanchit Gandhi}, \bibinfo{person}{Shaden Smith}, \bibinfo{person}{Stéphane Requena}, \bibinfo{person}{Suraj Patil}, \bibinfo{person}{Tim Dettmers}, \bibinfo{person}{Ahmed Baruwa}, \bibinfo{person}{Amanpreet Singh}, \bibinfo{person}{Anastasia Cheveleva}, \bibinfo{person}{Anne-Laure Ligozat}, \bibinfo{person}{Arjun Subramonian}, \bibinfo{person}{Aurélie Névéol}, \bibinfo{person}{Charles Lovering},
  \bibinfo{person}{Dan Garrette}, \bibinfo{person}{Deepak Tunuguntla}, \bibinfo{person}{Ehud Reiter}, \bibinfo{person}{Ekaterina Taktasheva}, \bibinfo{person}{Ekaterina Voloshina}, \bibinfo{person}{Eli Bogdanov}, \bibinfo{person}{Genta~Indra Winata}, \bibinfo{person}{Hailey Schoelkopf}, \bibinfo{person}{Jan-Christoph Kalo}, \bibinfo{person}{Jekaterina Novikova}, \bibinfo{person}{Jessica~Zosa Forde}, \bibinfo{person}{Jordan Clive}, \bibinfo{person}{Jungo Kasai}, \bibinfo{person}{Ken Kawamura}, \bibinfo{person}{Liam Hazan}, \bibinfo{person}{Marine Carpuat}, \bibinfo{person}{Miruna Clinciu}, \bibinfo{person}{Najoung Kim}, \bibinfo{person}{Newton Cheng}, \bibinfo{person}{Oleg Serikov}, \bibinfo{person}{Omer Antverg}, \bibinfo{person}{Oskar van~der Wal}, \bibinfo{person}{Rui Zhang}, \bibinfo{person}{Ruochen Zhang}, \bibinfo{person}{Sebastian Gehrmann}, \bibinfo{person}{Shachar Mirkin}, \bibinfo{person}{Shani Pais}, \bibinfo{person}{Tatiana Shavrina}, \bibinfo{person}{Thomas Scialom}, \bibinfo{person}{Tian Yun},
  \bibinfo{person}{Tomasz Limisiewicz}, \bibinfo{person}{Verena Rieser}, \bibinfo{person}{Vitaly Protasov}, \bibinfo{person}{Vladislav Mikhailov}, \bibinfo{person}{Yada Pruksachatkun}, \bibinfo{person}{Yonatan Belinkov}, \bibinfo{person}{Zachary Bamberger}, \bibinfo{person}{Zdeněk Kasner}, \bibinfo{person}{Alice Rueda}, \bibinfo{person}{Amanda Pestana}, \bibinfo{person}{Amir Feizpour}, \bibinfo{person}{Ammar Khan}, \bibinfo{person}{Amy Faranak}, \bibinfo{person}{Ana Santos}, \bibinfo{person}{Anthony Hevia}, \bibinfo{person}{Antigona Unldreaj}, \bibinfo{person}{Arash Aghagol}, \bibinfo{person}{Arezoo Abdollahi}, \bibinfo{person}{Aycha Tammour}, \bibinfo{person}{Azadeh HajiHosseini}, \bibinfo{person}{Bahareh Behroozi}, \bibinfo{person}{Benjamin Ajibade}, \bibinfo{person}{Bharat Saxena}, \bibinfo{person}{Carlos~Muñoz Ferrandis}, \bibinfo{person}{Daniel McDuff}, \bibinfo{person}{Danish Contractor}, \bibinfo{person}{David Lansky}, \bibinfo{person}{Davis David}, \bibinfo{person}{Douwe Kiela},
  \bibinfo{person}{Duong~A. Nguyen}, \bibinfo{person}{Edward Tan}, \bibinfo{person}{Emi Baylor}, \bibinfo{person}{Ezinwanne Ozoani}, \bibinfo{person}{Fatima Mirza}, \bibinfo{person}{Frankline Ononiwu}, \bibinfo{person}{Habib Rezanejad}, \bibinfo{person}{Hessie Jones}, \bibinfo{person}{Indrani Bhattacharya}, \bibinfo{person}{Irene Solaiman}, \bibinfo{person}{Irina Sedenko}, \bibinfo{person}{Isar Nejadgholi}, \bibinfo{person}{Jesse Passmore}, \bibinfo{person}{Josh Seltzer}, \bibinfo{person}{Julio~Bonis Sanz}, \bibinfo{person}{Livia Dutra}, \bibinfo{person}{Mairon Samagaio}, \bibinfo{person}{Maraim Elbadri}, \bibinfo{person}{Margot Mieskes}, \bibinfo{person}{Marissa Gerchick}, \bibinfo{person}{Martha Akinlolu}, \bibinfo{person}{Michael McKenna}, \bibinfo{person}{Mike Qiu}, \bibinfo{person}{Muhammed Ghauri}, \bibinfo{person}{Mykola Burynok}, \bibinfo{person}{Nafis Abrar}, \bibinfo{person}{Nazneen Rajani}, \bibinfo{person}{Nour Elkott}, \bibinfo{person}{Nour Fahmy}, \bibinfo{person}{Olanrewaju Samuel},
  \bibinfo{person}{Ran An}, \bibinfo{person}{Rasmus Kromann}, \bibinfo{person}{Ryan Hao}, \bibinfo{person}{Samira Alizadeh}, \bibinfo{person}{Sarmad Shubber}, \bibinfo{person}{Silas Wang}, \bibinfo{person}{Sourav Roy}, \bibinfo{person}{Sylvain Viguier}, \bibinfo{person}{Thanh Le}, \bibinfo{person}{Tobi Oyebade}, \bibinfo{person}{Trieu Le}, \bibinfo{person}{Yoyo Yang}, \bibinfo{person}{Zach Nguyen}, \bibinfo{person}{Abhinav~Ramesh Kashyap}, \bibinfo{person}{Alfredo Palasciano}, \bibinfo{person}{Alison Callahan}, \bibinfo{person}{Anima Shukla}, \bibinfo{person}{Antonio Miranda-Escalada}, \bibinfo{person}{Ayush Singh}, \bibinfo{person}{Benjamin Beilharz}, \bibinfo{person}{Bo Wang}, \bibinfo{person}{Caio Brito}, \bibinfo{person}{Chenxi Zhou}, \bibinfo{person}{Chirag Jain}, \bibinfo{person}{Chuxin Xu}, \bibinfo{person}{Clémentine Fourrier}, \bibinfo{person}{Daniel~León Periñán}, \bibinfo{person}{Daniel Molano}, \bibinfo{person}{Dian Yu}, \bibinfo{person}{Enrique Manjavacas}, \bibinfo{person}{Fabio Barth},
  \bibinfo{person}{Florian Fuhrimann}, \bibinfo{person}{Gabriel Altay}, \bibinfo{person}{Giyaseddin Bayrak}, \bibinfo{person}{Gully Burns}, \bibinfo{person}{Helena~U. Vrabec}, \bibinfo{person}{Imane Bello}, \bibinfo{person}{Ishani Dash}, \bibinfo{person}{Jihyun Kang}, \bibinfo{person}{John Giorgi}, \bibinfo{person}{Jonas Golde}, \bibinfo{person}{Jose~David Posada}, \bibinfo{person}{Karthik~Rangasai Sivaraman}, \bibinfo{person}{Lokesh Bulchandani}, \bibinfo{person}{Lu Liu}, \bibinfo{person}{Luisa Shinzato}, \bibinfo{person}{Madeleine~Hahn de Bykhovetz}, \bibinfo{person}{Maiko Takeuchi}, \bibinfo{person}{Marc Pàmies}, \bibinfo{person}{Maria~A Castillo}, \bibinfo{person}{Marianna Nezhurina}, \bibinfo{person}{Mario Sänger}, \bibinfo{person}{Matthias Samwald}, \bibinfo{person}{Michael Cullan}, \bibinfo{person}{Michael Weinberg}, \bibinfo{person}{Michiel~De Wolf}, \bibinfo{person}{Mina Mihaljcic}, \bibinfo{person}{Minna Liu}, \bibinfo{person}{Moritz Freidank}, \bibinfo{person}{Myungsun Kang},
  \bibinfo{person}{Natasha Seelam}, \bibinfo{person}{Nathan Dahlberg}, \bibinfo{person}{Nicholas~Michio Broad}, \bibinfo{person}{Nikolaus Muellner}, \bibinfo{person}{Pascale Fung}, \bibinfo{person}{Patrick Haller}, \bibinfo{person}{Ramya Chandrasekhar}, \bibinfo{person}{Renata Eisenberg}, \bibinfo{person}{Robert Martin}, \bibinfo{person}{Rodrigo Canalli}, \bibinfo{person}{Rosaline Su}, \bibinfo{person}{Ruisi Su}, \bibinfo{person}{Samuel Cahyawijaya}, \bibinfo{person}{Samuele Garda}, \bibinfo{person}{Shlok~S Deshmukh}, \bibinfo{person}{Shubhanshu Mishra}, \bibinfo{person}{Sid Kiblawi}, \bibinfo{person}{Simon Ott}, \bibinfo{person}{Sinee Sang-aroonsiri}, \bibinfo{person}{Srishti Kumar}, \bibinfo{person}{Stefan Schweter}, \bibinfo{person}{Sushil Bharati}, \bibinfo{person}{Tanmay Laud}, \bibinfo{person}{Théo Gigant}, \bibinfo{person}{Tomoya Kainuma}, \bibinfo{person}{Wojciech Kusa}, \bibinfo{person}{Yanis Labrak}, \bibinfo{person}{Yash~Shailesh Bajaj}, \bibinfo{person}{Yash Venkatraman}, \bibinfo{person}{Yifan
  Xu}, \bibinfo{person}{Yingxin Xu}, \bibinfo{person}{Yu Xu}, \bibinfo{person}{Zhe Tan}, \bibinfo{person}{Zhongli Xie}, \bibinfo{person}{Zifan Ye}, \bibinfo{person}{Mathilde Bras}, \bibinfo{person}{Younes Belkada}, {and} \bibinfo{person}{Thomas Wolf}.} \bibinfo{year}{2023}\natexlab{}.
\newblock \bibinfo{title}{BLOOM: A 176B-Parameter Open-Access Multilingual Language Model}.
\newblock
\newblock
\showeprint[arxiv]{2211.05100}~[cs.CL]


\bibitem[Wu et~al\mbox{.}(2019)]%
        {wu2019machine}
\bibfield{author}{\bibinfo{person}{Carole-Jean Wu}, \bibinfo{person}{David Brooks}, \bibinfo{person}{Kevin Chen}, \bibinfo{person}{Douglas Chen}, \bibinfo{person}{Sy Choudhury}, \bibinfo{person}{Marat Dukhan}, \bibinfo{person}{Kim Hazelwood}, \bibinfo{person}{Eldad Isaac}, \bibinfo{person}{Yangqing Jia}, \bibinfo{person}{Bill Jia}, {et~al\mbox{.}}} \bibinfo{year}{2019}\natexlab{}.
\newblock \showarticletitle{Machine learning at facebook: Understanding inference at the edge}. In \bibinfo{booktitle}{\emph{2019 IEEE international symposium on high performance computer architecture (HPCA)}}. IEEE, \bibinfo{pages}{331--344}.
\newblock


\bibitem[Xiao et~al\mbox{.}(2023)]%
        {xiao2023smoothquant}
\bibfield{author}{\bibinfo{person}{Guangxuan Xiao}, \bibinfo{person}{Ji Lin}, \bibinfo{person}{Mickael Seznec}, \bibinfo{person}{Hao Wu}, \bibinfo{person}{Julien Demouth}, {and} \bibinfo{person}{Song Han}.} \bibinfo{year}{2023}\natexlab{}.
\newblock \showarticletitle{Smoothquant: Accurate and efficient post-training quantization for large language models}. In \bibinfo{booktitle}{\emph{International Conference on Machine Learning}}. PMLR, \bibinfo{pages}{38087--38099}.
\newblock


\bibitem[Xu et~al\mbox{.}(2022)]%
        {xu2022distributed}
\bibfield{author}{\bibinfo{person}{Yuzhe Xu}, \bibinfo{person}{Thaha Mohammed}, \bibinfo{person}{Mario Di~Francesco}, {and} \bibinfo{person}{Carlo Fischione}.} \bibinfo{year}{2022}\natexlab{}.
\newblock \showarticletitle{Distributed Assignment With Load Balancing for DNN Inference at the Edge}.
\newblock \bibinfo{journal}{\emph{IEEE Internet of Things Journal}} \bibinfo{volume}{10}, \bibinfo{number}{2} (\bibinfo{year}{2022}), \bibinfo{pages}{1053--1065}.
\newblock


\bibitem[Yao et~al\mbox{.}(2022)]%
        {yao2022zeroquant}
\bibfield{author}{\bibinfo{person}{Zhewei Yao}, \bibinfo{person}{Reza Yazdani~Aminabadi}, \bibinfo{person}{Minjia Zhang}, \bibinfo{person}{Xiaoxia Wu}, \bibinfo{person}{Conglong Li}, {and} \bibinfo{person}{Yuxiong He}.} \bibinfo{year}{2022}\natexlab{}.
\newblock \showarticletitle{Zeroquant: Efficient and affordable post-training quantization for large-scale transformers}.
\newblock \bibinfo{journal}{\emph{Advances in Neural Information Processing Systems}}  \bibinfo{volume}{35} (\bibinfo{year}{2022}), \bibinfo{pages}{27168--27183}.
\newblock


\bibitem[Zeng et~al\mbox{.}(2021)]%
        {coedge}
\bibfield{author}{\bibinfo{person}{Liekang Zeng}, \bibinfo{person}{Xu Chen}, \bibinfo{person}{Zhi Zhou}, \bibinfo{person}{Lei Yang}, {and} \bibinfo{person}{Junshan Zhang}.} \bibinfo{year}{2021}\natexlab{}.
\newblock \showarticletitle{CoEdge: Cooperative DNN Inference With Adaptive Workload Partitioning Over Heterogeneous Edge Devices}.
\newblock \bibinfo{journal}{\emph{IEEE/ACM Trans. Netw.}} \bibinfo{volume}{29}, \bibinfo{number}{2} (\bibinfo{date}{apr} \bibinfo{year}{2021}), \bibinfo{pages}{595–608}.
\newblock
\showISSN{1063-6692}
\urldef\tempurl%
\url{https://doi.org/10.1109/TNET.2020.3042320}
\showDOI{\tempurl}


\bibitem[Zhang et~al\mbox{.}(2019)]%
        {zhang2019deep}
\bibfield{author}{\bibinfo{person}{Chaoyun Zhang}, \bibinfo{person}{Paul Patras}, {and} \bibinfo{person}{Hamed Haddadi}.} \bibinfo{year}{2019}\natexlab{}.
\newblock \showarticletitle{Deep learning in mobile and wireless networking: A survey}.
\newblock \bibinfo{journal}{\emph{IEEE Communications surveys \& tutorials}} \bibinfo{volume}{21}, \bibinfo{number}{3} (\bibinfo{year}{2019}), \bibinfo{pages}{2224--2287}.
\newblock


\bibitem[Zhang et~al\mbox{.}(2022)]%
        {zhang2022opt}
\bibfield{author}{\bibinfo{person}{Susan Zhang}, \bibinfo{person}{Stephen Roller}, \bibinfo{person}{Naman Goyal}, \bibinfo{person}{Mikel Artetxe}, \bibinfo{person}{Moya Chen}, \bibinfo{person}{Shuohui Chen}, \bibinfo{person}{Christopher Dewan}, \bibinfo{person}{Mona Diab}, \bibinfo{person}{Xian Li}, \bibinfo{person}{Xi~Victoria Lin}, \bibinfo{person}{Todor Mihaylov}, \bibinfo{person}{Myle Ott}, \bibinfo{person}{Sam Shleifer}, \bibinfo{person}{Kurt Shuster}, \bibinfo{person}{Daniel Simig}, \bibinfo{person}{Punit~Singh Koura}, \bibinfo{person}{Anjali Sridhar}, \bibinfo{person}{Tianlu Wang}, {and} \bibinfo{person}{Luke Zettlemoyer}.} \bibinfo{year}{2022}\natexlab{}.
\newblock \bibinfo{title}{OPT: Open Pre-trained Transformer Language Models}.
\newblock
\newblock
\showeprint[arxiv]{2205.01068}~[cs.CL]


\bibitem[Zhao et~al\mbox{.}(2022)]%
        {zhao2022survey}
\bibfield{author}{\bibinfo{person}{Tianming Zhao}, \bibinfo{person}{Yucheng Xie}, \bibinfo{person}{Yan Wang}, \bibinfo{person}{Jerry Cheng}, \bibinfo{person}{Xiaonan Guo}, \bibinfo{person}{Bin Hu}, {and} \bibinfo{person}{Yingying Chen}.} \bibinfo{year}{2022}\natexlab{}.
\newblock \showarticletitle{A survey of deep learning on mobile devices: Applications, optimizations, challenges, and research opportunities}.
\newblock \bibinfo{journal}{\emph{Proc. IEEE}} \bibinfo{volume}{110}, \bibinfo{number}{3} (\bibinfo{year}{2022}), \bibinfo{pages}{334--354}.
\newblock


\bibitem[Zhou et~al\mbox{.}(2019)]%
        {adaptive_para_exec}
\bibfield{author}{\bibinfo{person}{Li Zhou}, \bibinfo{person}{Mohammad~Hossein Samavatian}, \bibinfo{person}{Anys Bacha}, \bibinfo{person}{Saikat Majumdar}, {and} \bibinfo{person}{Radu Teodorescu}.} \bibinfo{year}{2019}\natexlab{}.
\newblock \showarticletitle{Adaptive Parallel Execution of Deep Neural Networks on Heterogeneous Edge Devices}. In \bibinfo{booktitle}{\emph{Proceedings of the 4th ACM/IEEE Symposium on Edge Computing}} (Arlington, Virginia) \emph{(\bibinfo{series}{SEC '19})}. \bibinfo{publisher}{Association for Computing Machinery}, \bibinfo{address}{New York, NY, USA}, \bibinfo{pages}{195–208}.
\newblock
\showISBNx{9781450367332}
\urldef\tempurl%
\url{https://doi.org/10.1145/3318216.3363312}
\showDOI{\tempurl}


\end{thebibliography}

\appendix

\end{document}